\documentclass{article} 
\usepackage{colm2024_conference}

\usepackage{booktabs}
\usepackage{graphicx}
\usepackage{enumitem}
\usepackage{wrapfig}
\usepackage{algorithm}
\usepackage{algpseudocode}
\usepackage{natbib}
\usepackage{makecell}
\usepackage{booktabs}
\usepackage{array}
\usepackage{amsmath}
\usepackage{amssymb}
\usepackage{amsfonts}
\usepackage{multirow}
\usepackage{verbatim}
\usepackage{caption}
\usepackage{longtable}
\usepackage{supertabular}
\usepackage{CJKutf8}
\usepackage[utf8]{inputenc} 
\usepackage[T1]{fontenc}
\usepackage[french,vietnamese,mongolian,greek,english]{babel}
\usepackage{pifont}

\usepackage{enumitem}
\usepackage{tablefootnote}
\usepackage{xspace}
\usepackage{textcomp}
\usepackage{makecell}
\usepackage{lscape} 
\usepackage{siunitx}
\usepackage{listings}
\usepackage{xcolor}

\definecolor{dt}{gray}{0.7}
\usepackage{pifont}       
\usepackage{bbding}       
\usepackage{fontawesome}

\usepackage{scrextend}

\usepackage{tgpagella}
\usepackage{latexsym}
\usepackage[T1]{fontenc}
\usepackage[utf8]{inputenc}
\usepackage{microtype}
\definecolor{mydarkblue}{rgb}{0,0.08,0.45}
\definecolor{citecolor}{HTML}{0071BC}
\usepackage{url}            
\usepackage{nicefrac}       
\usepackage{changepage}
\usepackage{xargs}          
\usepackage{wrapfig,lipsum,booktabs}
\usepackage{longtable}
\usepackage{subcaption}
\usepackage{endnotes}

\usepackage{pgfplots}
\usetikzlibrary{pgfplots.groupplots}
\pgfplotsset{compat=1.3}
\usepackage{tikz}
\usetikzlibrary{patterns}

\usepackage[most]{tcolorbox}

\usepackage[capitalize,noabbrev]{cleveref}
\crefname{section}{Section}{\S\S}
\Crefname{section}{Section}{\S\S}
\crefname{table}{Table}{Tables}
\crefname{figure}{Figure}{Figures}
\crefname{algorithm}{Algorithm}{}
\crefname{equation}{eq.}{}
\crefname{appendix}{Appendix}{}
\crefformat{section}{Section #2#1#3}
\usepackage{multicol}
\usepackage{tcolorbox}
\usepackage{titlesec}
\titleformat*{\section}{\large\bfseries}

\title{Qwen2.5-VL Technical Report}

\author{
\bf Qwen Team, Alibaba Group}

\begin{document}

\maketitle

\begin{abstract}
We introduce Qwen2.5-VL, the latest flagship model of Qwen vision-language series, which demonstrates significant advancements in both foundational capabilities and innovative functionalities. Qwen2.5-VL achieves a major leap forward in understanding and interacting with the world through enhanced visual recognition, precise object localization, robust document parsing, and long-video comprehension.
A standout feature of Qwen2.5-VL is its ability to localize objects using bounding boxes or points accurately. It provides robust structured data extraction from invoices, forms, and tables, as well as detailed analysis of charts, diagrams, and layouts. To handle complex inputs, Qwen2.5-VL introduces dynamic resolution processing and absolute time encoding, enabling it to process images of varying sizes and videos of extended durations (up to hours) with second-level event localization. This allows the model to natively perceive spatial scales and temporal dynamics without relying on traditional normalization techniques.
By training a native dynamic-resolution Vision Transformer (ViT) from scratch and incorporating Window Attention, we have significantly reduced computational overhead while maintaining native resolution. As a result, Qwen2.5-VL excels not only in static image and document understanding but also as an interactive visual agent capable of reasoning, tool usage, and task execution in real-world scenarios such as operating computers and mobile devices. The model achieves strong generalization across domains without requiring task-specific fine-tuning.
Qwen2.5-VL is available in three sizes, addressing diverse use cases from edge AI to high-performance computing. The flagship Qwen2.5-VL-72B model matches state-of-the-art models like GPT-4o and Claude 3.5 Sonnet, particularly excelling in document and diagram understanding. The smaller Qwen2.5-VL-7B and Qwen2.5-VL-3B models outperform comparable competitors, offering strong capabilities even in resource-constrained environments. Additionally, Qwen2.5-VL maintains robust linguistic performance, preserving the core language competencies of the Qwen2.5 LLM.
\end{abstract}
\begin{figure*}[ht]
\centering
\includegraphics[width= 1\linewidth]{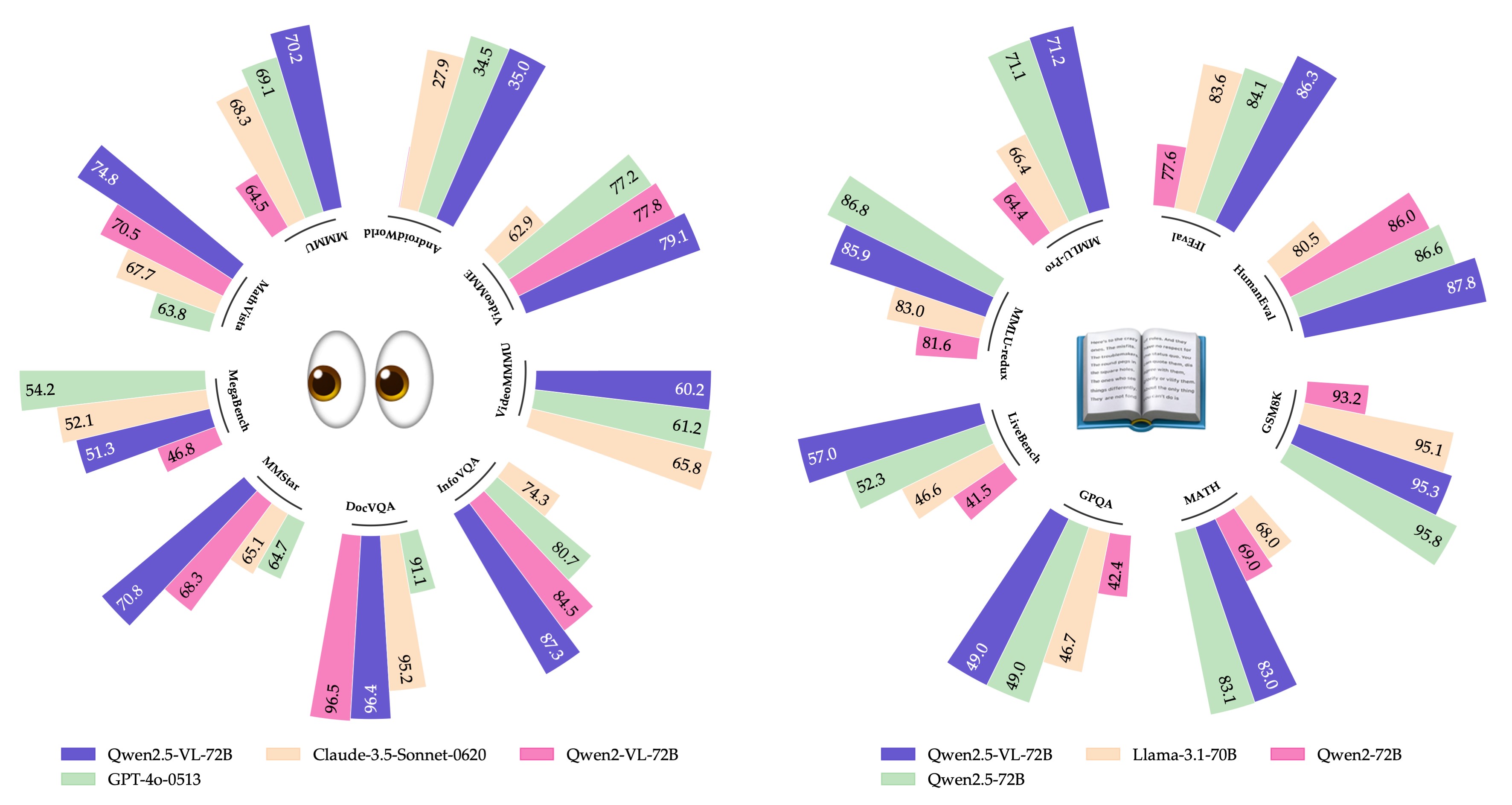}
\end{figure*}

\section{Introduction}
Large vision-language models ( LVLMs )~\citep{gpt4o,sonnet3_5,team2023gemini,wang2024qwen2} represent a pivotal breakthrough in artificial intelligence, signaling a transformative approach to multimodal understanding and interaction. By seamlessly integrating visual perception with natural language processing, these advanced models are fundamentally reshaping how machines interpret and analyze complex information across diverse domains. Despite significant advancements in multimodal large language models, the current capabilities of these models can be likened to the middle layer of a sandwich cookie—competent across various tasks but falling short of exceptional performance. Fine-grained visual tasks form the foundational layer of this analogy. In this iteration of Qwen2.5-VL, we are committed to exploring fine-grained perception capabilities, aiming to establish a robust foundation for LVLMs and create an agentic amplifier for real-world applications. The top layer of this framework is multi-modal reasoning, which is enhanced by leveraging the latest Qwen2.5 LLM and employing multi-modal QA data construction.

A spectrum of works have promoted the development of multimodal large models, characterized by architectural design, visual input processing, and data curation. One of the primary drivers of progress in LVLMs is the continuous innovation in architecture. The studies presented in~\citep{flamingo, blip, blip2, llava, llava1.5, wang2024emu3, zhang2024internlm,wang2023internimage} have incrementally shaped the current paradigm, which typically consists of a visual encoder, a cross-modal projector, and LLM. Fine-grained perception models have emerged as another crucial area. Models like ~\citep{xiao2023florence,grounding_dino,ren2024grounding, ferretv2,zhang2024omg,kosmos2,deitke2024molmo} have pushed the boundaries of what is possible in terms of detailed visual understanding. The architectures of Omni~\citep{li2024baichuan,li2025baichuan,ye2024x} and MoE~\citep{riquelme2021scaling,lee2024moai,li2024uni,li2024aria,wu2024deepseek} also inspire the future evolution of LVLMs. Enhancements in visual encoders~\citep{internvl, liu2024points,liang2025global} and resolution scaling~\citep{monkey,mplug-owl2,Otterhd} have played a pivotal role in improving the quality of practical visual understanding. Curating data with more diverse scenarios and higher-quality is an essential step in training advanced LVLMs. The efforts proposed in~\citep{guo2024mammoth,chen2024expanding, liu2024mminstruct, chen2024allava, tong2024cambrian, li2024llava} are highly valuable contributions to this endeavor. 

However, despite their remarkable progress, vision-language models currently face developmental bottlenecks, including computational complexity,  limited contextual understanding, poor fine-grained visual perception, and inconsistent performance across varied sequence length.

In this report, we introduce the latest work Qwen2.5-VL, which continues the open-source philosophy of the Qwen series, achieving and even surpassing top-tier closed-source models on various benchmarks. Technically, our contributions are four-folds: (1) We implement window attention in the visual encoder to optimize inference efficiency; (2) We introduce dynamic FPS sampling, extending dynamic resolution to the temporal dimension and enabling comprehensive video understanding across varied sampling rates; (3) We upgrade MRoPE in the temporal domain by aligning to absolute time, thereby facilitating more sophisticated temporal sequence learning; (4) We make significant efforts in curating high-quality data for both pre-training and supervised fine-tuning, further scaling the pre-training corpus from 1.2 trillion tokens to 4.1 trillion tokens.

The sparkling characteristics of Qwen2.5-VL are as follows:
\begin{itemize}

 \item \textbf{Powerful document parsing capabilities:} Qwen2.5-VL upgrades text recognition to omni-document parsing, excelling in processing multi-scene, multilingual, and various built-in (handwriting, tables, charts, chemical formulas, and music sheets) documents.
 
 \item \textbf{Precise object grounding across formats:} Qwen2.5-VL unlocks improved accuracy in detecting, pointing, and counting objects, accommodating absolute coordinate and JSON formats for advanced spatial reasoning.
 
 \item \textbf{Ultra-long video understanding and fine-grained video grounding:} Our model extends native dynamic resolution to the temporal dimension, enhancing the ability to understand videos lasting hours while extracting event segments in seconds.
 
 \item \textbf{Enhanced agent Functionality for computer and mobile devices:} Leverage advanced grounding, reasoning, and decision-making abilities, boosting the model with superior agent functionality on smartphones and computers.
 
\end{itemize}

\section{Approach}

In this section, we first outline the architectural updates of the Qwen2.5-VL series models and provide an overview of the data and training details. 

\begin{figure*}[t]
\centering
\includegraphics[width= 1\linewidth]{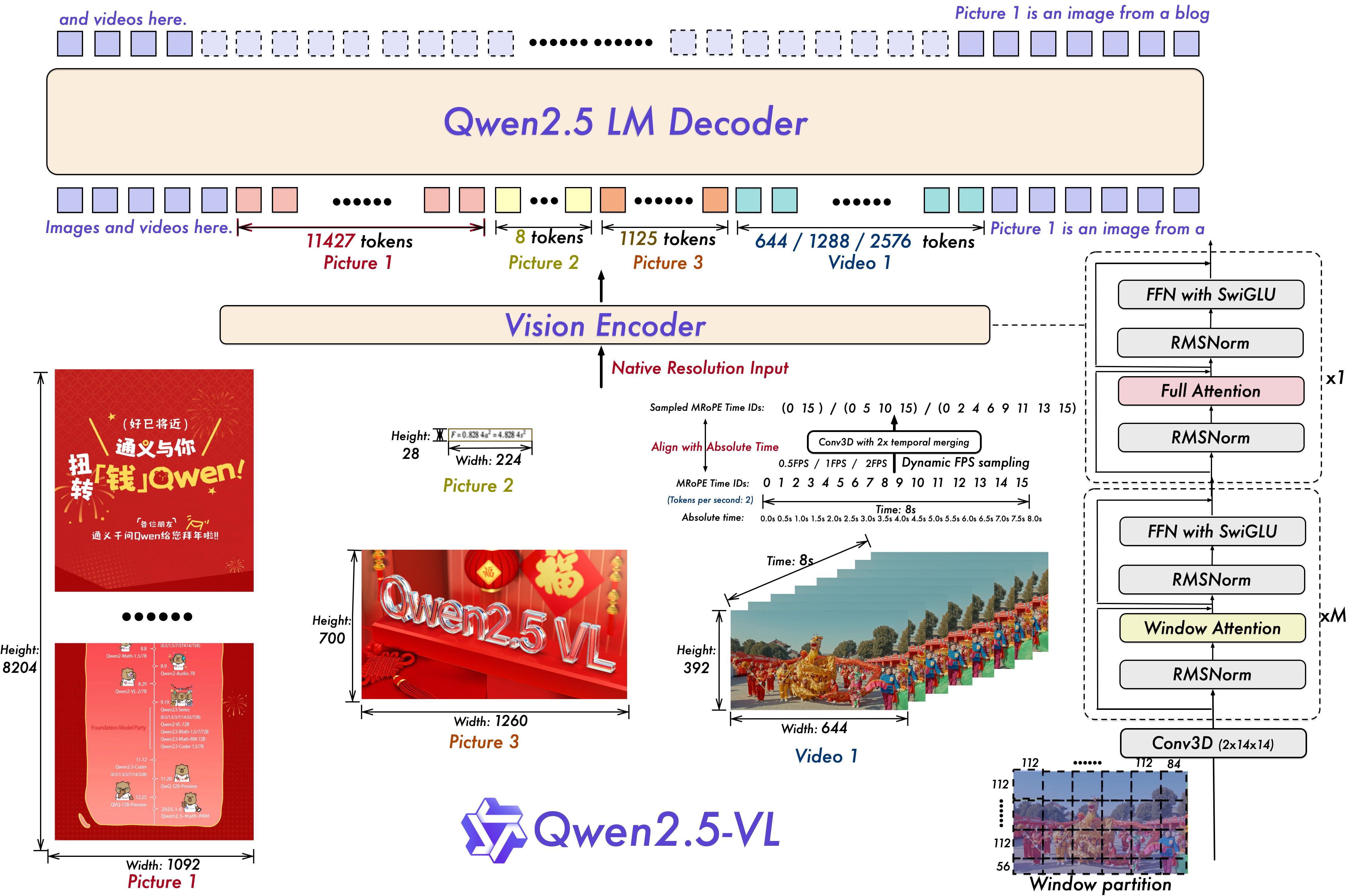}
   \caption{The Qwen2.5-VL framework demonstrates the integration of a vision encoder and a language model decoder to process multimodal inputs, including images and videos. The vision encoder is designed to handle inputs at their native resolution and supports dynamic FPS sampling. Images of varying sizes and video frames with different FPS rates are dynamically mapped to token sequences of varying lengths. Notably, MRoPE aligns time IDs with absolute time along the temporal dimension, enabling the model to better comprehend temporal dynamics, such as the pace of events and precise moment localization. The processed visual data is subsequently fed into the Qwen2.5 LM Decoder. We have re-engineered the vision transformer (ViT) architecture, incorporating advanced components such as FFN with SwiGLU activation, RMSNorm for normalization, and window-based attention mechanisms to enhance performance and efficiency.
}
\label{fig:arc}
\end{figure*}

\subsection{Model Architecture}
The overall model architecture of Qwen2.5-VL consists of three components:

\textbf{Large Language Model}: The Qwen2.5-VL series adopts large language models as its foundational component. The model is initialized with pre-trained weights from the Qwen2.5 LLM. To better meet the demands of multimodal understanding, we have modified the 1D RoPE (Rotary Position Embedding) to our Multimodal Rotary Position Embedding Aligned to Absolute Time.

\textbf{Vision Encoder}: The vision encoder of Qwen2.5-VL employs a redesigned Vision Transformer (ViT) architecture. Structurally, we incorporate 2D-RoPE and window attention to support native input resolutions while accelerating the computation of the entire visual encoder. During both training and inference, the height and width of the input images are resized to multiples of 28 before being fed into the ViT. The vision encoder processes images by splitting them into patches with a stride of 14, generating a set of image features. We provide a more detailed introduction to the vision encoder in Section~\ref{sec:VisionEncoder}.

\textbf{MLP-based Vision-Language Merger}: To address the efficiency challenges posed by long sequences of image features, we adopt a simple yet effective approach to compress the feature sequences before feeding them into the large language model (LLM). Specifically, instead of directly using the raw patch features extracted by the Vision Transformer (ViT), we first group spatially adjacent sets of four patch features. These grouped features are then concatenated and passed through a two-layer multi-layer perceptron (MLP) to project them into a dimension that aligns with the text embeddings used in the LLM. This method not only reduces computational costs but also provides a flexible way to dynamically compress image feature sequences of varying lengths.

In Table~\ref{table:config}, the architecture and configuration of Qwen2.5-VL are detailed.

\begin{table}[htbp]
\centering
\begin{tabular}{lccc}
\toprule
\textbf{Configuration} & \textbf{Qwen2.5-VL-3B} & \textbf{Qwen2.5-VL-7B} & \textbf{Qwen2.5-VL-72B} \\ \midrule
\multicolumn{4}{c}{\textbf{Vision Transformer (ViT) }} \\ \midrule
Hidden Size & 1280 & 1280 & 1280 \\
\# Layers & 32 & 32 & 32 \\
\# Num Heads & 16 & 16 & 16 \\
Intermediate Size & 3456 & 3456 & 3456 \\
Patch Size & 14 & 14 & 14 \\
Window Size & 112 & 112 & 112 \\
Full Attention Block Indexes & \{7, 15, 23, 31\} & \{7, 15, 23, 31\} & \{7, 15, 23, 31\} \\ \midrule
\multicolumn{4}{c}{\textbf{Vision-Language Merger}} \\  \midrule
In Channel & 1280 & 1280 & 1280 \\
Out Channel & 2048 & 3584 & 8192 \\ \midrule
\multicolumn{4}{c}{\textbf{Large Language Model (LLM) }} \\ \midrule
Hidden Size & 2048 & 3,584 & 8192 \\
\# Layers & 36 & 28 & 80 \\
\# KV Heads & 2 & 4 & 8 \\
Head Size & 128 & 128 & 128 \\
Intermediate Size & 4864 & 18944 & 29568 \\
Embedding Tying & \ding{51} & \ding{55} & \ding{55} \\
Vocabulary Size & 151646 & 151646 & 151646 \\
\# Trained Tokens & 4.1T & 4.1T & 4.1T \\ 

\bottomrule
\end{tabular}
\caption{Configuration of Qwen2.5-VL.}
\label{table:config}
\end{table}

\subsubsection{Fast and Efficient Vision Encoder}
\label{sec:VisionEncoder}

The vision encoder plays a pivotal role in multimodal large language models (MLLMs). To address the challenges posed by computational load imbalances during training and inference due to native resolution inputs, we have redesigned the Vision Transformer (ViT) architecture. A key issue arises from the quadratic computational complexity associated with processing images of varying sizes. To mitigate this, we introduce windowed attention in most layers, which ensures that computational cost scales linearly with the number of patches rather than quadratically. In our architecture, only four layers employ full self-attention, while the remaining layers utilize windowed attention with a maximum window size of 112×112 (corresponding to 8×8 patches). Regions smaller than 112×112 are processed without padding, preserving their original resolution. This design allows the model to operate natively at the input resolution, avoiding unnecessary scaling or distortion.

For positional encoding, we adopt 2D Rotary Positional Embedding (RoPE) to effectively capture spatial relationships in 2D space. Furthermore, to better handle video inputs, we extend our approach to 3D patch partitioning. Specifically, we use 14×14 image patches as the basic unit, consistent with traditional ViTs for static images. For video data, two consecutive frames are grouped together, significantly reducing the number of tokens fed into the language model. This design not only maintains compatibility with existing architectures but also enhances efficiency when processing sequential video data.

To streamline the overall network structure, we align the ViT architecture more closely with the design principles of large language models (LLMs). Specifically, we adopt RMSNorm~\citep{rmsnorm} for normalization and SwiGLU~\citep{glu} as the activation function. These choices enhance both computational efficiency and compatibility between the vision and language components of the model.

In terms of training, we train the redesigned ViT from scratch. The training process consists of several stages, including CLIP pre-training, vision-language alignment, and end-to-end fine-tuning. To ensure robustness across varying input resolutions, we employ dynamic sampling at native resolutions during training. Images are randomly sampled according to their original aspect ratios, enabling the model to generalize effectively to inputs of diverse resolutions. This approach not only improves the model's adaptability but also ensures stable and efficient training across different sizes of visual data. 

\subsubsection{Native Dynamic Resolution and Frame Rate}

Qwen2.5-VL introduces advancements in both spatial and temporal dimensions to handle diverse multimodal inputs effectively.

In the spatial domain, Qwen2.5-VL dynamically converts images of varying sizes into sequences of tokens with corresponding lengths. Unlike traditional approaches that normalize coordinates, our model directly uses the actual dimensions of the input image to represent bounding boxes, points, and other spatial features. This allows the model to learn scale information inherently, improving its ability to process images across different resolutions.

For video inputs, Qwen2.5-VL incorporates dynamic frame rate (FPS) training and absolute time encoding. By adapting to variable frame rates, the model can better capture the temporal dynamics of video content.
Unlike other approaches that incorporate textual timestamps or utilize additional heads to enable temporal grounding, we introduce a novel and efficient strategy that aligns MRoPE IDs directly with the timestamps. This approach allows the model to understand the tempo of time through the intervals between temporal dimension IDs, without necessitating any additional computational overhead.

\subsubsection{Multimodal Rotary Position Embedding Aligned to Absolute Time}

Positional embeddings are crucial for modeling sequential data in both vision and language modalities. Building upon the Multimodal Rotary Position Embedding (MRoPE) introduced in Qwen2-VL, we extend its capabilities to better handle temporal information in videos.

The MRoPE in Qwen2-VL decomposes the position embedding into three distinct components: temporal, height, and width to effectively model multimodal inputs. For textual inputs, all three components use identical position IDs, making MRoPE functionally equivalent to traditional 1D RoPE~\citep{rope}. For images, the temporal ID remains constant across visual tokens, while unique IDs are assigned to the height and width components based on each token's spatial position within the image. When processing videos, which are treated as sequences of frames, the temporal ID increments for each frame, while the height and width components follow the same assignment pattern as for static images.

However, in Qwen2-VL, the temporal position IDs in MRoPE were tied to the number of input frames, which did not account for the speed of content changes or the absolute timing of events within the video. To address this limitation, Qwen2.5-VL introduces a key improvement: aligning the temporal component of MRoPE with absolute time. As shown in Figure~\ref{fig:arc}, by leveraging the intervals between temporal IDs, the model is able to learn consistent temporal alignment across videos with different FPS sampling rates.

\subsection{Pre-Training}
In this section, we first describe the construction of the pre-training dataset, followed by an overview of the overall training pipeline and configuration.
\subsubsection{Pre-Training Data}
Compared to Qwen2-VL, we have significantly expanded the volume of our pre-training data, increasing it from 1.2 trillion tokens to approximately 4 trillion tokens. Our pre-training dataset was constructed through a combination of methods, including cleaning raw web data, synthesizing data, etc. The dataset encompasses a wide variety of multimodal data, such as image captions, interleaved image-text data, optical character recognition (OCR) data, visual knowledge (e.g., celebrity, landmark, flora, and fauna identification), multi-modal academic questions, localization data, document parsing data, video descriptions, video localization, and agent-based interaction data. Throughout the training process, we carefully adjusted the composition and proportions of these data types at different stages to optimize learning outcomes.

\paragraph{Interleaved Image-Text Data}
Interleaved image-text data is essential for multimodal learning, offering three key benefits: (1) enabling in-context learning with simultaneous visual and textual cues~\citep{flamingo}, (2) maintaining strong text-only capabilities when images are missing~\citep{lin2024vila}, and (3) containing a wide range of general information. However, much of the available interleaved data lacks meaningful text-image associations and is often noisy, limiting its usefulness for complex reasoning and creative generation.

To address these challenges, we developed a pipeline for scoring and cleaning data, ensuring only high-quality, relevant interleaved data is used. Our process involves two steps: standard data cleaning~\citep{li2024omnicorpus} followed by a four-stage scoring system using an internal evaluation model. The scoring criteria include: (1) text-only quality, (2) image-text relevance, (3) image-text complementarity, and (4) information density balance. This meticulous approach improves the model’s ability to perform complex reasoning and generate coherent multimodal content.

The following is a description of these image-text scoring criteria:

Image-text Relevance: A higher score indicates a stronger connection between the image and text, where the image meaningfully supplements, explains or expands on the text rather than just decorating it.

Information Complementarity: A higher score reflects greater complementary information between the image and text. Each should provide unique details that together create a complete narrative.

Balance of Information Density: A higher score means a more balanced distribution of information between the image and text, avoiding excessive text or image information, and ensuring an appropriate balance between the two.

\paragraph{Grounding Data with Absolute Position Coordinates}
We adopt native resolution training with the aim of achieving a more accurate perception of the world. In contrast, relative coordinates fail to effectively represent the original size and position of objects within images. To address this limitation, Qwen2.5-VL uses coordinate values based on the actual dimensions of the input images during training to represent bounding boxes and points. This approach ensures that the model can better capture the real-world scale and spatial relationships of objects, leading to improved performance in tasks such as object detection and localization.

To improve the generalizability of grounding capabilities, we have developed a comprehensive dataset encompassing bounding boxes and points with referring expressions, leveraging both publicly available datasets and proprietary data. 
Our methodology involves synthesizing data into various formats, including XML, JSON, and custom formats, employing techniques such as copy-paste augmentation~\citep{ghiasi2021simple} and synthesis with off-the-shelf models such as Grounding DINO~\citep{grounding_dino} and SAM~\citep{kirillov2023segment}. This approach facilitates a more robust evaluation and advancement of grounding abilities.

To enhance the model's performance on open-vocabulary detection, we expanded the training dataset to include over 10,000 object categories. Additionally, to improve the model's effectiveness in extreme object detection scenarios, we synthesized non-existent object categories within the queries and constructed image data containing multiple instances for each object.

To ensure superior point-based object grounding capabilities, we have constructed a comprehensive pointing dataset comprising both publicly available and synthetic data. Specifically, the data source includes public pointing and counting data from PixMo~\citep{deitke2024molmo}, publicly accessible object grounding data (from both object detection and instance segmentation tasks), and data synthesized by an automated pipeline for generating precise pointing data towards certain image details.

\paragraph{Document Omni-Parsing Data}
To train Qwen2.5-VL, we synthesized a large corpus of document data. Traditional methods for parsing document content typically rely on separate models to handle layout analysis, text extraction, chart interpretation, and illustration processing. In contrast, Qwen2.5-VL is designed to empower a general-purpose model with comprehensive capabilities for parsing, understanding, and converting document formats. Specifically, we incorporated a diverse array of elements into the documents, such as tables, charts, equations, natural or synthetic images, music sheets, and chemical formulas. These elements were uniformly formatted in HTML, which integrates layout box information and descriptions of illustrations into HTML tag structures. We also enriched the document layouts according to typical reading sequences and included the coordinates corresponding to each module, such as paragraphs and charts, in the HTML-based ground truth. This innovative approach allows the complete information of any document, including its layout, text, charts, and illustrations, to be represented in a standardized and unified manner. As a result, Qwen2.5-VL achieves seamless integration of multimodal document elements, thereby facilitating more efficient and accurate document understanding and transformation.

Below is the QwenVL HTML format:

\begin{tcolorbox}[colback=black!5!white,colframe=black!75!black,title=QwenVL HTML Format]
<html><body> \\
\textcolor{blue}{\# paragraph} \\
<p data-bbox="\textcolor{red}{x1 y1 x2 y2}"> \textcolor{red}{content} </p> \\
\textcolor{blue}{\# table} \\
<style>\textcolor{red}{table\{id\} style}</style><table data-bbox="\textcolor{red}{x1 y1 x2 y2}" class="table\textcolor{red}{\{id\}}"> \textcolor{red}{table content} </table> \\
\textcolor{blue}{\# chart} \\
<div class="chart" data-bbox="\textcolor{red}{x1 y1 x2 y2}"> <img data-bbox="\textcolor{red}{x1 y1 x2 y2}" /><table> \textcolor{red}{chart content} </table></div> \\
\textcolor{blue}{\# formula} \\
<div class="formula" data-bbox="\textcolor{red}{x1 y1 x2 y2}"> <img data-bbox="\textcolor{red}{x1 y1 x2 y2}" /> <div> \textcolor{red}{formula content} </div></div>\\
\textcolor{blue}{\# image caption} \\
<div class="image caption" data-bbox="\textcolor{red}{x1 y1 x2 y2}"> <img data-bbox="\textcolor{red}{x1 y1 x2 y2}" /><p> \textcolor{red}{image caption} </p></div>\\
\textcolor{blue}{\# image ocr} \\
<div class="image ocr" data-bbox="\textcolor{red}{x1 y1 x2 y2}"> <img data-bbox="\textcolor{red}{x1 y1 x2 y2}" /><p> \textcolor{red}{image ocr} </p></div>\\
\textcolor{blue}{\# music sheet} \\
<div class="music sheet" format="abc notation" data-bbox="\textcolor{red}{x1 y1 x2 y2}"> <img data-bbox="\textcolor{red}{x1 y1 x2 y2}" /> <div> \textcolor{red}{music sheet content} </div></div> \\
\textcolor{blue}{\# chemical formula content} \\
<div class="chemical formula" format="smile" data-bbox="\textcolor{red}{x1 y1 x2 y2}"> <img data-bbox="\textcolor{red}{x1 y1 x2 y2}" /> <div> \textcolor{red}{chemical formula content} </div></div> \\
</html></body>
\end{tcolorbox}

This format ensures that all document elements are represented in a structured and accessible manner, enabling efficient processing and understanding by Qwen2.5-VL.

\paragraph{OCR Data} Data from different sources are gathered and curated to enhance the OCR performance, including synthetic data, open-sourced data and in-house collected data. Synthetic data is generated through a visual text generation engine to produce high-quality text images in the wild. To support a wider range of languages and enhance multilingual capabilities, we have incorporated a large-scale multilingual OCR dataset. This dataset includes support for diverse languages such as French, German, Italian, Spanish, Portuguese, Arabic, Russian, Japanese, Korean, and Vietnamese. The dataset is carefully curated to ensure diversity and quality, utilizing both high-quality synthetic images and real-world natural scene images. This combination ensures robust performance across various linguistic contexts and improves the model’s adaptability to different text appearances and environmental conditions. For chart-type data, we synthesized 1 million samples using visualization libraries including matplotlib, seaborn, and plotly, encompassing chart categories such as bar charts, relational diagrams, and heatmaps. Regarding tabular data, we processed 6 million real-world samples through an offline end-to-end table recognition model, subsequently filtering out low-confidence tables, overlapping tables, and tables with insufficient cell density.

\paragraph{Video Data}
To ensure enhanced robustness in understanding video data with varying frames per second (FPS), we dynamically sampled FPS during training to achieve a more evenly distributed representation of FPS within the training dataset. Additionally, for videos exceeding half an hour in length, we specifically constructed a set of long video captions by synthesizing multi-frame captions through a targeted synthesis pipeline. Regarding video grounding data, we formulated timestamps in both second-based formats and hour-minute-second-frame (hmsf) formats, ensuring that the model can accurately understand and output time in various formats.

\paragraph{Agent Data}

We enhance the perception and decision-making abilities to build the agent capabilities of Qwen2.5-VL. For perception, we collect screenshots on mobile, web, and desktop platforms. A synthetic data engine is used to generate screenshot captions and UI element grounding annotations. The caption task helps Qwen2.5-VL understand the graphic interface, while the grounding task helps it align the appearance and function of elements. For decision-making, we first unify the operations across mobile, web, and desktop platforms into a function call format with a shared action space. A set of annotated multi-step trajectories collected from open-source data and synthesized by agent framework~\citep{wang2025mobile, wang2024mobile2, wang2024mobile} on virtual environments are reformatted into a function format. We further generate a reasoning process for each step through human and model annotators~\citep{xu2024aguvis}. Specifically, given a ground-truth operation, we highlight it on the screenshot. Then, we provide the global query, along with screenshots from before and after this operation, to the annotators and require them to write reasoning content to explain the intention behind this operation. A model-based filter is used to screen out low-quality reasoning content. Such reasoning content prevents Qwen2.5-VL from overfitting to the ground-truth operations and makes it more robust in real-world scenarios.

\begin{table}[h!]
\small
\centering
\begin{tabular}{lccc}
\toprule
\textbf{Stages}        & \textbf{Visual Pre-Training}       & \textbf{Multimodal Pre-Training}       & \textbf{Long-Context Pre-Training}       \\ \midrule
Data   &   \makecell{Image Caption \\ Knowledge \\ OCR}     & \makecell{+ \\ Pure text \\ Interleaved Data \\ VQA, Video \\ Grounding, Agent}   &     \makecell{+ \\ 
 Long Video\\ Long Agent \\ Long Document}    \\ 
\midrule
Tokens&  1.5T  &    2T   &    0.6T   \\ 
\midrule
Sequence length   &  8192  &    8192   &    32768   \\ 
\midrule
Training   &  ViT  &    ViT \& LLM   &    ViT \& LLM   \\ 
\bottomrule
\end{tabular}
\caption{Training data volume and composition across different stages.}
\label{tab:pretrainingdata}
\end{table}

\subsubsection{Training Recipe}
We trained a Vision Transformer (ViT) from scratch using DataComp~\citep{datacomp} and some in-house datasets as the initialization for the vision encoder, while leveraging the pre-trained Qwen2.5 large language model (LLM)~\citep{qwen2.5} as the initialization for the LLM component. As shown in Table \ref{tab:pretrainingdata}, the pre-training process is divided into three distinct phases, each employing different data configurations and training strategies to progressively enhance the model's capabilities.

In the first phase, only the Vision Transformer (ViT) is trained to improve its alignment with the language model, laying a solid foundation for multimodal understanding. The primary data sources during this phase include image captions, visual knowledge, and OCR data. These datasets are carefully selected to foster ViT's ability to extract meaningful visual representations that can be effectively integrated with textual information.

In the second phase, all model parameters are unfrozen, and the model is trained on a diverse set of multimodal image data to enhance its capacity to process complex visual information. This phase introduces more intricate and reasoning-intensive datasets, such as interleaved data, multi-task learning datasets, visual question answering (VQA), multimodal mathematics, agent-based tasks, video understanding, and pure-text datasets. These datasets strengthen the model's ability to establish deeper connections between visual and linguistic modalities, enabling it to handle increasingly sophisticated tasks.

In the third phase, to further enhance the model's reasoning capabilities over longer sequences, video, and agent-based data are incorporated, alongside an increase in sequence length. This allows the model to tackle more advanced and intricate multimodal tasks with greater precision. By extending the sequence length, the model gains the ability to process extended contexts, which is particularly beneficial for tasks requiring long-range dependencies and complex reasoning.

To address the challenges posed by varying image sizes and text lengths, which can lead to imbalanced computational loads during training, we adopted a strategy to optimize training efficiency. The primary computational costs arise from the LLM and the vision encoder. Given that the vision encoder has relatively fewer parameters and that we introduced window attention to further reduce its computational demands, we focused on balancing the computational load of the LLM across different GPUs. Specifically, we dynamically packed data samples based on their corresponding input sequence lengths to the LLM, ensuring consistent computational loads. In the first and second phases, data were uniformly packed to a sequence length of 8,192, while in the third phase, the sequence length was increased to 32,768 to accommodate the model's enhanced capacity for handling longer sequences.

\subsection{Post-training}

The post-training alignment framework of Qwen2.5-VL employs a dual-stage optimization paradigm comprising Supervised Fine-Tuning (SFT) and Direct Preference Optimization (DPO)~\citep{DBLP:conf/nips/RafailovSMMEF23}. This hierarchical alignment strategy synergizes parameter-efficient domain adaptation with human preference distillation, addressing both representational grounding and behavioral refinement through distinct optimization objectives.

Supervised Fine-Tuning (SFT) aims to bridge the gap between pretrained representations and downstream task requirements through targeted instruction optimization. During this phase, we employ the ChatML format~\citep{chatml} to structure instruction-following data, deliberately diverging from the pretraining data schema while maintaining architectural consistency with Qwen2-VL~\citep{Qwen2-VL}. This format transition enables three critical adaptations: 1) Explicit dialogue role tagging for multimodal turn-taking, 2) Structured injection of visual embeddings alongside textual instructions, and 3) Preservation of cross-modal positional relationships through format-aware packing. By exposing the model to curated multimodal instruction-response pairs under this enhanced schema, SFT enables efficient knowledge transfer while maintaining the integrity of pre-trained features. 

\subsubsection{Instruction Data}

The Supervised Fine-Tuning (SFT) phase employs a meticulously curated dataset designed to enhance the model's instruction-following capabilities across diverse modalities. This dataset comprises approximately 2 million entries, evenly distributed between pure text data (50\%) and multimodal data (50\%), which includes image-text and video-text combinations. The inclusion of multimodal data enables the model to process complex inputs effectively. Notably, although pure text and multimodal entries are equally represented, multimodal entries consume significantly more tokens and computational resources during training due to the embedded visual and temporal information. The dataset is primarily composed of Chinese and English data, with supplementary multilingual entries to support broader linguistic diversity.

The dataset is structured to reflect varying levels of dialogue complexity, including both single-turn and multi-turn interactions. These interactions are further contextualized by scenarios ranging from single-image inputs to multi-image sequences, thereby simulating realistic conversational dynamics. The query sources are primarily drawn from open-source repositories, with additional contributions from curated purchased datasets and online query data. This combination ensures broad coverage and enhances the representativeness of the dataset.

To address a wide range of application scenarios, the dataset includes specialized subsets for General Visual Question Answering (VQA), image captioning, mathematical problem-solving, coding tasks, and security-related queries. Additionally, dedicated datasets for Document and Optical Character Recognition (Doc and OCR), Grounding, Video Analysis, and Agent Interactions are constructed to enhance domain-specific proficiency. Detailed information regarding the data can be found in the relevant sections of the paper. This structured and diverse composition ensures that the SFT phase effectively aligns pre-trained representations with the nuanced demands of downstream multimodal tasks, fostering robust and contextually aware model performance.

\subsubsection{Data Filtering Pipeline}

The quality of training data is a critical factor influencing the performance of vision-language models. Open-source and synthetic datasets typically exhibit significant variability, often containing noisy, redundant, or low-quality samples. Therefore, rigorous data cleaning and filtering processes are essential to address these issues. Low-quality data can lead to suboptimal alignment between pretrained representations and downstream task requirements, thereby diminishing the model's ability to effectively handle complex multimodal tasks. Consequently, ensuring high-quality data is paramount for achieving robust and reliable model performance.

To address these challenges, we implement a two-stage data filtering pipeline designed to systematically enhance the quality of the Supervised Fine-Tuning (SFT) dataset. This pipeline comprises the following stages:

\paragraph{Stage 1: Domain-Specific Categorization}

In the initial stage, we employ \textit{Qwen2-VL-Instag}, a specialized classification model derived from Qwen2-VL-72B, to perform hierarchical categorization of question-answer (QA) pairs. This model organizes QA pairs into eight primary domains, such as \textit{Coding} and \textit{Planning}, which are further divided into 30 fine-grained subcategories. For example, the primary domain \textit{Coding} is subdivided into subcategories including \textit{Code\_Debugging}, \textit{Code\_Generation}, \textit{Code\_Translation}, and \textit{Code\_Understanding}. This hierarchical structure facilitates domain-aware and subdomain-aware filtering strategies, enabling the pipeline to optimize data-cleaning processes tailored to each category's specific characteristics. Consequently, this enhances the quality and relevance of the supervised fine-tuning (SFT) dataset.

\paragraph{Stage 2: Domain-Tailored Filtering}

The second stage involves domain-tailored filtering, which integrates both rule-based and model-based approaches to comprehensively enhance data quality. Given the diverse nature of domains such as Document Processing, Optical Character Recognition (OCR), and Visual Grounding, each may necessitate unique filtering strategies. Below, we provide an overview of the general filtering strategies applied across these domains.

\textbf{Rule-Based Filtering} employs predefined heuristics to eliminate low-quality or problematic entries. Specifically, for datasets related to Document Processing, OCR, and Visual Grounding tasks, repetitive patterns are identified and removed to prevent distortion of the model's learning process and ensure optimal performance. Additionally, entries containing incomplete, truncated, or improperly formatted responses—common in synthetic datasets and multimodal contexts—are excluded. To maintain relevance and uphold ethical standards, queries and answers that are unrelated or could potentially lead to harmful outputs are also discarded. This structured approach ensures that the dataset adheres to ethical guidelines and meets task-specific requirements.

\textbf{Model-Based Filtering} further refines the dataset by leveraging reward models trained on the Qwen2.5-VL series. These models evaluate multimodal QA pairs across multiple dimensions. Queries are assessed for complexity and relevance, retaining only those examples that are appropriately challenging and contextually pertinent. Answers are evaluated based on correctness, completeness, clarity, relevance to the query, and helpfulness. In visual-grounded tasks, particular attention is given to verifying the accurate interpretation and utilization of visual information. This multi-dimensional scoring ensures that only high-quality data progresses to the SFT phase.

\subsubsection{Rejection Sampling for Enhanced Reasoning}

To complement our structured data filtering pipeline, we employ rejection sampling as a strategy to refine the dataset and enhance the reasoning capabilities of the vision-language model (VLM). This approach is particularly critical for tasks requiring complex inference, such as mathematical problem-solving, code generation, and domain-specific visual question answering (VQA). Prior research has shown that incorporating Chain-of-Thought (CoT) \cite{DBLP:journals/corr/abs-2201-11903} reasoning significantly improves a model's inferential performance.~\citep{DBLP:journals/corr/abs-2412-19437} Our post-training experiments confirm this, underscoring the importance of structured reasoning processes for achieving high-quality outcomes.

The rejection sampling process begins with datasets enriched with ground truth annotations. These datasets are carefully curated to include tasks that demand multi-step reasoning, such as mathematical problem-solving, code generation, and domain-specific VQA. Using an intermediate version of the Qwen2.5-VL model, we evaluate the generated responses against the ground truth. Only samples where the model's output matches the expected answers are retained, ensuring the dataset consists solely of high-quality, accurate examples.

To further improve data quality, we apply additional constraints to filter out undesirable outputs. Specifically, we exclude responses that exhibit code-switching, excessive length, or repetitive patterns. These criteria ensure clarity and coherence in the CoT reasoning process, which is crucial for downstream applications.

A key challenge in applying CoT reasoning to vision-language models is their reliance on both textual and visual modalities. Intermediate reasoning steps may fail to adequately integrate visual information, either by ignoring relevant visual cues or misinterpreting them. To address this, we have developed rule-based and model-driven filtering strategies to validate the accuracy of intermediate reasoning steps. These mechanisms ensure that each step in the CoT process effectively integrates visual and textual modalities. Despite these efforts, achieving optimal modality alignment remains an ongoing challenge that requires further advancements.

The data generated through rejection sampling significantly enhances the model's reasoning proficiency. By iteratively refining the dataset and removing low-quality or erroneous samples, we enable the model to learn from high-fidelity examples that emphasize accurate and coherent reasoning. This methodology not only strengthens the model's ability to handle complex tasks but also lays the groundwork for future improvements in vision-language modeling.

\subsubsection{Training Recipe}
The post-training process for Qwen2.5-VL consists of two phases: Supervised Fine-Tuning (SFT) and Direct Preference Optimization (DPO), both with the Vision Transformer (ViT) parameters frozen. In the SFT phase, the model is fine-tuned on diverse multimodal data, including image-text pairs, video, and pure text, sourced from general VQA, Rejection Sampling, and specialized datasets such as Document and OCR, Grounding, Video, and Agent-related tasks. The DPO phase focuses exclusively on image-text and pure text data, utilizing preference data to align the model with human preferences, with each sample processed only once to ensure efficient optimization. This streamlined process enhances the model’s cross-modal reasoning and task-specific performance while maintaining alignment with user intent.

\section{Experiments}
In this section, we first introduce the overall model and compare it with the current state-of-the-art (SoTA) models. Then, we evaluate the model's performance across various sub-capabilities.

\subsection{Comparison with the SOTA Models}

\begin{table}[h]
\centering
\caption{\textbf{Performance of Qwen2.5-VL and State-of-the-art.}}
\label{tab:sota_results}
\setlength{\tabcolsep}{3.0pt}
\scalebox{0.63}{
\begin{tabular}{@{}lcccccccc@{}}
\toprule
\textbf{Datasets}  & \makecell{\textbf{Previous} \\ \textbf{Open-source SoTA}} & \makecell{\textbf{Claude-3.5} \\ \textbf{Sonnet-0620}} & \makecell{\textbf{GPT-4o} \\ \textbf{0513}} & \makecell{\textbf{InternVL2.5} \\\textbf{78B}} & \makecell{\textbf{Qwen2-VL} \\ \textbf{72B}} & \makecell{\textbf{Qwen2.5-VL} \\ \textbf{72B}} & \makecell{\textbf{Qwen2.5-VL} \\ \textbf{7B}}  & \makecell{\textbf{Qwen2.5-VL} \\ \textbf{3B}}  \\ 
\midrule
\multicolumn{9}{c}{\textit{College-level Problems}} \\
\midrule
MMMU$_{\text{val}}$~\citep{yue2023mmmu} & 70.1~\cite{chen2024expanding} & 68.3 & 69.1 & 70.1 & 64.5 & \textbf{70.2} & 58.6 & 53.1  \\ 
MMMU-Pro$_{\text{overall}}$~\citep{mmmupro} & 48.6~\cite{chen2024expanding} & 51.5 & \textbf{51.9} & 48.6 & 46.2 & 51.1 & 38.3 & 31.56  \\ 
\midrule
\multicolumn{9}{c}{\textit{Math}} \\
\midrule
MathVista$_{\text{mini}}$~\citep{mathvista}   & 72.3~\cite{chen2024expanding} & 67.7 & 63.8 & 72.3 & 70.5 & \textbf{74.8} & 68.2 & 62.3  \\ 
MATH-Vision$_{\text{full}}$~\citep{mathvision}   & 32.2~\cite{chen2024expanding} & - & 30.4 & 32.2 & 25.9 & \textbf{38.1} & 25.1 & 21.2  \\    
MathVerse$_{\text{mini}}$~\citep{zhang2024mathverse}  & 51.7~\cite{chen2024expanding} & - & 50.2 & 51.7 & - & \textbf{57.6} & 49.2 & 47.6  \\ 
\midrule
\multicolumn{9}{c}{\textit{General Visual Question Answering}} \\
\midrule
MegaBench~\citep{chen2024mega}  & 47.4~\cite{minimax2025minimax01scalingfoundationmodels} & 52.1 & \textbf{54.2} & 45.6 & 46.8 & 51.3 & 36.8 & 28.9  \\ 
MMBench-EN$_{\text{test}}$~\citep{MMBench}  & 88.3~\cite{chen2024expanding} & 82.6 & 83.4 & 88.3 & 86.9 & \textbf{88.6} & 83.5 & 79.1  \\ 
MMBench-CN$_{\text{test}}$~\citep{MMBench}  & 88.5~\cite{chen2024expanding} & 83.5 & 82.1 & \textbf{88.5} & 86.7 & 87.9 & 83.4 & 78.1  \\ 
MMBench-V1.1-EN$_{\text{test}}$~\citep{MMBench}  & 87.4~\cite{chen2024expanding} & 80.9 & 83.1 & 87.4 & 86.1 & \textbf{88.4} & 82.6 & 77.4  \\ 
MMStar~\citep{chen2024we}     & 69.5~\cite{chen2024expanding} & 65.1 & 64.7 & 69.5 & 68.3 & \textbf{70.8} & 63.9 & 55.9  \\ 
MME$_{\text{sum}}$~\citep{fu2023mme}  & \textbf{2494}~\cite{chen2024expanding} & 1920 & 2328 & \textbf{2494} & 2483 & 2448 & 2347 & 2157  \\ 
MuirBench~\citep{wang2024muirbench}  & 63.5~\cite{chen2024expanding} & - & 68.0 & 63.5 & - & \textbf{70.7} & 59.6 & 47.7  \\ 
BLINK$_{\text{val}}$~\citep{fu2024blink}  & 63.8~\cite{chen2024expanding} & - & \textbf{68.}0 & 63.8 & - & 64.4 & 56.4 & 47.6  \\ 
CRPE$_{\text{relation}}$~\citep{wang2024allseeing_v2}  & 78.8~\cite{chen2024expanding} & - & 76.6 & 78.8 & - & \textbf{79.2} & 76.4 & 73.6  \\ 
HallBench$_{\text{avg}}$~\citep{guan2023hallusionbench}  & \textbf{58.1}~\cite{wang2024qwen2} & 55.5 & 55.0 & 57.4 & \textbf{58.1} & 55.2 & 52.9 & 46.3  \\ 
MTVQA~\citep{tang2024mtvqa}  & \textbf{31.9}~\cite{chen2024expanding} & 25.7 & 27.8 & 31.9 & 30.9 & 31.7 & 29.2 & 24.8  \\ 
RealWorldQA$_{\text{avg}}$~\citep{grok15}  & 78.7~\cite{chen2024expanding} & 60.1 & 75.4 & \textbf{78.7} & 77.8 & 75.7 & 68.5 & 65.4  \\ 
MME-RealWorld$_{\text{en}}$~\citep{mme-realworld}  & 62.9~\cite{chen2024expanding} & 51.6 & 45.2 & 62.9 & - & \textbf{63.2} & 57.4 & 53.1  \\ 
MMVet$_{\text{turbo}}$~\citep{yu2024mm}  & 74.0~\cite{wang2024qwen2} & 70.1 & 69.1 & 72.3 & 74.0 & \textbf{76.2} & 67.1 & 61.8  \\ 
MM-MT-Bench~\citep{agrawal2024pixtral}  & 7.4~\cite{agrawal2024pixtral} & 7.5 & \textbf{7.72} & - & 6.59 & 7.6 & 6.3 & 5.7  \\ 

\bottomrule
\end{tabular}
}
\end{table}

The experimental section evaluates the performance of Qwen2.5-VL across a variety of datasets, comparing it with state-of-the-art models such as Claude-3.5-Sonnet-0620~\citep{sonnet3_5}, GPT-4o-0513~\citep{gpt4o}, InternVL2.5~\citep{chen2024expanding}, and different sizes of Qwen2-VL~\citep{Qwen2-VL}. In college-level problems, Qwen2.5-VL-72B achieves a score of 70.2 on MMMU~\citep{yue2023mmmu}. For MMMU-Pro~\citep{mmmupro}, Qwen2.5-VL-72B scores 51.1, surpassing the previous open-source state-of-the-art models and achieving performance comparable to GPT-4o.

In math-related tasks, Qwen2.5-VL-72B demonstrates strong capabilities. On MathVista~\citep{mathvista}, it achieves a score of 74.8, outperforming the previous open-source state-of-the-art score of 72.3. For MATH-Vision~\citep{mathvision}, Qwen2.5-VL-72B scores 38.1, while MathVerse~\citep{zhang2024mathverse} achieves 57.6, both showing competitive results compared to other leading models.

For general visual question answering, Qwen2.5-VL-72B excels across multiple benchmarks. On MMbench-EN~\citep{MMBench}, it achieves a score of 88.6, slightly surpassing the previous best score of 88.3. The model also performs well in MuirBench~\citep{wang2024muirbench} with a score of 70.7 and BLINK~\citep{fu2024blink} with 64.4. In the multilingual capability evaluation of MTVQA~\citep{tang2024mtvqa}, Qwen2.5-VL-72B achieves a score of 31.7, showcasing its powerful multilingual text recognition abilities. In subjective evaluations such as MMVet~\citep{yu2024mm} and MM-MT-Bench~\citep{agrawal2024pixtral}, Qwen2.5-VL-72B scores 76.2 and 7.6, respectively, demonstrating excellent natural conversational experience and user satisfaction.

\subsection{Performance on Pure Text Tasks}
To critically evaluate the performance of instruction-tuned models on pure text tasks, as illustrated in Table~\ref{puretext}, we selected several representative benchmarks to assess the model's capabilities across a variety of domains, including general tasks~\citep{mmlupro,mmluredux,livebench}, mathematics and science tasks~\citep{gpqa,math,gsm8k}, coding tasks~\citep{humaneval,multiple}, and alignment task~\citep{ifeval}. We compared Qwen2.5-VL with several large language models (LLMs) of similar size. The results demonstrate that Qwen2.5-VL not only achieves state-of-the-art (SoTA) performance on multimodal tasks but also exhibits leading performance on pure text tasks, showcasing its versatility and robustness across diverse evaluation criteria.

\begin{table}[tbp]

\centering

\caption{\textbf{Performance on pure text tasks of the 70B+ Instruct models and Qwen2.5-VL.}}

\label{tab:70b_instruct}
\small
\setlength{\tabcolsep}{2.6pt}

\begin{tabular}{@{}lccccc@{}}

\toprule

\textbf{Datasets}  & \textbf{Llama-3.1-70B} & \textbf{Llama-3.1-405B} & \textbf{Qwen2-72B} & \textbf{Qwen2.5-72B} & \textbf{Qwen2.5-VL-72B} \\

\midrule

\multicolumn{6}{c}{\textit{General Tasks}} \\
\midrule

MMLU-Pro & 66.4 & \textbf{73.3} & 64.4 & 71.1 & 71.2 \\

MMLU-redux  & 83.0 & 86.2 & 81.6 & \textbf{86.8} & 85.9 \\

LiveBench-0831 & 46.6 & 53.2 & 41.5 & 52.3 & \textbf{57.0} \\

\midrule
\multicolumn{6}{c}{\textit{Mathematics \& Science Tasks}} \\
\midrule

GPQA  & 46.7 & \textbf{51.1} & 42.4 & 49.0 & 49.0\\

MATH  & 68.0 & 73.8 & 69.0 & \textbf{83.1} & 83.0\\

GSM8K  & 95.1 & \textbf{96.8} & 93.2 & 95.8 & 95.3\\

\midrule
\multicolumn{6}{c}{\textit{Coding Tasks}} \\
\midrule

HumanEval  & 80.5 & \textbf{89.0} & 86.0 & 86.6 & 87.8\\

MultiPL-E  & 68.2 & 73.5 & 69.2 & 75.1 & \textbf{79.5} \\

\midrule
\multicolumn{6}{c}{\textit{Alignment Tasks}} \\
\midrule

IFEval  & 83.6 & 86.0 & 77.6 & 84.1 & \textbf{86.3}\\

\bottomrule

\end{tabular}
\label{puretext}
\end{table}

\subsection{Quantitative Results}

\subsubsection{General Visual Question Answering}
To comprehensively evaluate the model's capabilities in general visual question answering (VQA) and dialogue, we conducted extensive experiments across a diverse range of datasets. As illustrated in Table~\ref{tab:sota_results}, Qwen2.5-VL demonstrates state-of-the-art performance in various VQA tasks, subjective evaluations, multilingual scenarios, and multi-image questions. Specifically, it excels on benchmark datasets such as MMBench series~\citep{MMBench}, MMStar~\citep{chen2024we}, MME~\citep{fu2023mme}, MuirBench~\citep{wang2024muirbench}, BLINK\citep{fu2024blink}, CRPE~\citep{wang2024allseeing_v2}, HallBench~\citep{guan2023hallusionbench}, MTVQA~\citep{tang2024mtvqa}, MME-RealWorld~\citep{mme-realworld}, MMVet~\citep{yu2024mm}, and MM-MT-Bench~\citep{agrawal2024pixtral}.

In the domain of visual detail comprehension and reasoning, Qwen2.5-VL-72B achieves an accuracy of 88.4\% on the MMBench-EN-V1.1 dataset, surpassing previous state-of-the-art models such as InternVL2.5 (78B) and Claude-3.5 Sonnet-0620. Similarly, on the MMStar dataset, Qwen2.5-VL attains a score of 70.8\%, outperforming other leading models in this benchmark. These results underscore the model's robustness and adaptability across diverse linguistic contexts.

Furthermore, in high-resolution real-world scenarios, specifically on the MME-RealWorld benchmark, Qwen2.5-VL demonstrates state-of-the-art performance with a score of 63.2, showcasing its broad adaptability to realistic environments. Additionally, in multi-image understanding tasks evaluated on the MuirBench dataset, Qwen2.5-VL achieves a leading score of 70.7, further highlighting its superior generalization capabilities. Collectively, these results illustrate the strong versatility and effectiveness of Qwen2.5-VL in addressing general-purpose visual question answering (VQA) tasks across various scenarios.

Notably, even the smaller-scale versions of Qwen2.5-VL, specifically Qwen2.5-VL-7B and Qwen2.5-VL-3B, exhibit highly competitive performance. For instance, on the MMStar dataset, Qwen2.5-VL-7B achieves 63.9\%, while Qwen2.5-VL-3B scores 55.9\%. This demonstrates that Qwen2.5-VL's architecture is not only powerful but also scalable, maintaining strong performance even with fewer parameters.

\subsubsection{Document Understanding and OCR}

We evaluated our models across a diverse range of OCR, chart, and document understanding benchmarks. Table~\ref{tab:ocr_results} demonstrates the performance comparison between Qwen2.5-VL models and top-tier models on following OCR-related benchmarks: AI2D~\citep{kembhavi2016diagram}, TextVQA~\citep{textvqa}, DocVQA~\citep{docvqa}, InfoVQA~\citep{Mathew2021InfographicVQA}, ChartQA~\citep{masry2022chartqa}, CharXiv~\citep{wang2024charxiv}, SEED-Bench-2-Plus~\citep{li2024seed2plus}, OCRBench~\citep{liu2024ocrbenchhiddenmysteryocr},  OCRBench\_v2~\citep{fu2024ocrbenchv2improvedbenchmark}, CC-OCR~\citep{yang2024ccocrcomprehensivechallengingocr}, OmniDocBench~\citep{ouyang2024omnidocbenchbenchmarkingdiversepdf}, VCR~\citep{zhang2024vcr}.

For OCR-related parsing benchmarks on element parsing for multi-scene, multilingual, and various built-in (handwriting, tables, charts, chemical formulas, and mathematical expressions) documents, as CC-OCR and OmniDocBench, Qwen2.5-VL-72B model sets the new state-of-the-art due to curated training data and excellent capability of LLM models.

For OCR-related understanding benchmarks for scene text, chart, diagram and document, Qwen2.5-VL models achieve impressive performance with good understanding abilities. Notably, on composite OCR-related understanding benchmarks as OCRBench, InfoVQA which focusing on infographics, and SEED-Bench-2-Plus covering text-rich scenarios including charts, maps, and webs, Qwen2.5-VL-72B
achieves remarkable results, significantly outperforming strong competitors such as InternVL2.5-78B.

Furthermore, for OCR-related comprehensive benchmarks as OCRBench\_v2 including a wide range of OCR-related parsing and understanding tasks, top performance is also achieved by Qwen2.5-VL models, largely exceeding best model Gemini 1.5-Pro by 9.6\% and 20.6\% for English and Chinese track respectively. 

\begin{table}[h]

\centering
\caption{\textbf{Performance of Qwen2.5-VL and other models on OCR, chart, and document understanding benchmarks.}}
\label{tab:ocr_results}
\setlength{\tabcolsep}{3.0pt}
\scalebox{0.83}{
\begin{tabular}{@{}lcccccccc@{}}
\toprule
\textbf{Datasets}      & \makecell{\textbf{Claude-3.5} \\ \textbf{Sonnet}} & \makecell{\textbf{Gemini 1.5} \\ \textbf{Pro}} & \makecell{\textbf{GPT} \\ \textbf{4o}} & \makecell{\textbf{InternVL2.5} \\\textbf{78B}}&  \makecell{\textbf{Qwen2.5-VL} \\ \textbf{72B}} & \makecell{\textbf{Qwen2.5-VL} \\ \textbf{7B}}  & \makecell{\textbf{Qwen2.5-VL} \\ \textbf{3B}}  \\ 

\midrule
\multicolumn{8}{c}{\textit{OCR-related Parsing Tasks}} \\
\midrule

CC-OCR     & 62.5 & 73.0   & 66.9 & 64.7 & \textbf{79.8} & 77.8 & 74.5  \\ 
OmniDocBench$_{\text{edit en/zh}}$$\downarrow$ & 0.330/0.381 & 0.230/\textbf{0.281} & 0.265/0.435 & 0.275/0.324 & \textbf{0.226}/0.324 & 0.308/0.398 & 0.409/0.543  \\ 

\midrule
\multicolumn{8}{c}{\textit{OCR-related Understanding Tasks}} \\
\midrule
AI2D$_{\text{w. M.}}$ & 81.2 & 88.4 & 84.6 & \textbf{89.1} & 88.7 & 83.9 & 81.6  \\ 
TextVQA$_{\text{val}}$ & 76.5  & 78.8 & 77.4 & 83.4 & 83.5 & \textbf{84.9} & 79.3  \\ 
DocVQA$_{\text{test}}$ & 95.2   & 93.1 & 91.1 &95.1 &   \textbf{96.4} & 95.7 & 93.9     \\     
InfoVQA$_{\text{test}}$ & 74.3 & 81.0 & 80.7 & 84.1 & \textbf{87.3} & 82.6 & 77.1  \\ 
ChartQA$_{\text{test Avg.}}$ & \textbf{90.8} & 87.2 & 86.7 & 88.3 & 89.5 & 87.3 & 84.0  \\ 
CharXiv$_{\text{RQ/DQ}}$ & \textbf{60.2}/84.3  & 43.3/72.0 & 47.1/84.5 & 42.4/82.3 & 49.7/\textbf{87.4} & 42.5/73.9 & 31.3/58.6  \\ 
SEED-Bench-2-Plus  & 71.7 & 70.8 & 72.0 & 71.3 & \textbf{73.0} & 70.4 & 67.6  \\ 
OCRBench  & 788 & 754 & 736 & 854 &  \textbf{885} & 864 & 797  \\
VCR$_{\text{En-Hard-EM}}$  & 41.7 & 28.1 & 73.2 & - & 79.8 & \textbf{80.5} & 37.5  \\
\midrule
\multicolumn{8}{c}{\textit{OCR-related Comprehensive Tasks}} \\
\midrule
OCRBench\_v2$_{\text{en/zh}}$ & 45.2/39.6 & 51.9/43.1    & 46.5/32.2   & 49.8/52.1 & \textbf{61.5}/\textbf{63.7} & 56.3/57.2 & 54.3/52.1  \\ 
\bottomrule
\end{tabular}
}
\end{table}

\subsubsection{Spatial Understanding}

Understanding spatial relationships is crucial for developing AI models that can interpret and interact with the world as humans do. In Large Vision-Language Models, visual grounding allows for the precise localization and identification of specific objects, regions, or elements within an image based on natural language queries or descriptions. This capability transcends traditional object detection by establishing a semantic relationship between visual content and linguistic context, facilitating more nuanced and contextually aware visual reasoning. 
We evaluated Qwen2.5-VL's grounding capabilities on the referring expression comprehension benchmarks~\citep{refcoco, refcocog}, object detection in the wild ~\citep{li2022grounded}, self-curated point grounding benchmark, and CountBench~\citep{paiss2023teaching}.

We compare Qwen2.5-VL's visual grounding capabilities with other leading LVLMs including Gemini, Grounding-DINO~\citep{grounding_dino}, Molmo~\citep{deitke2024molmo}, and InternVL2.5.

Qwen2.5-VL achieves leading performance across different benchmarks from box-grounding, and point-grounding to counting.
By equipping Qwen2.5-VL with both box and point-grounding capability, it is able to understand, locate, and reason on the very details of certain parts of an image. For open-vocabulary object detection, Qwen2.5-VL achieves a good performance of 43.1 mAP on ODinW-13, surpassing most LVLMs and quickly narrowing the gap between generalist models and specialist models. In addition, Qwen2.5-VL unlocks the point-based grounding ability so that it could precisely locate the very details of a certain object, which was difficult to represent by a bounding box in the past. Qwen2.5-VL's counting ability also makes great progress, achieving a leading accuracy of 93.6 on CountBench with Qwen2.5-VL-72B using a ``detect then count''-style prompt.

\begin{table}[h]
\centering
\caption{\textbf{Performance of Qwen2.5-VL and other models on grounding.}}
\label{tab:grounding_results}
\scalebox{0.93}{
\setlength{\tabcolsep}{3.0pt}

\begin{tabular}{@{}lccccccc@{}}
\toprule
\textbf{Datasets}       & \makecell{\textbf{Gemini 1.5} \\ \textbf{Pro}} & \makecell{\textbf{Grounding} \\ \textbf{DINO}} & \makecell{\textbf{Molmo} \\ \textbf{72B}} & \makecell{\textbf{InternVL2.5} \\ \textbf{78B}} & \makecell{\textbf{Qwen2.5-VL} \\ \textbf{72B}} & \makecell{\textbf{Qwen2.5-VL} \\ \textbf{7B}} & \makecell{\textbf{Qwen2.5-VL} \\ \textbf{3B}} \\ 
\midrule
Refcoco$_{val}$ & 73.2   & 90.6 & - & 93.7 & 92.7 & 90.0 & 89.1 \\ 
Refcoco$_{testA}$ & 72.9    & 93.2 & - & 95.6 & 94.6 & 92.5 & 91.7 \\  
Refcoco$_{testB}$ & 74.6   & 88.2 & - & 92.5 & 89.7 & 85.4 & 84.0 \\  
\midrule
Refcoco+$_{val}$ & 62.5   & 88.2 & -  & 90.4 & 88.9 & 84.2 & 82.4 \\  
Refcoco+$_{testA}$ & 63.9   & 89.0 & - & 94.7 & 92.2 & 89.1 & 88.0 \\  
Refcoco+$_{testB}$ & 65.0    & 75.9 & - & 86.9 & 83.7 & 76.9 & 74.1 \\ 
\midrule
Refcocog$_{val}$ & 75.2   & 86.1 & - & 92.7 & 89.9 & 87.2 & 85.2 \\  
Refcocog$_{test}$ & 76.2    & 87.0 & - & 92.2 & 90.3 & 87.2 & 85.7 \\   
\midrule

ODinW  & 36.7 & 55.0 & - & 31.7 & 43.1 & 37.3 & 37.5 \\
\midrule
PointGrounding  & - & - & 69.2 & - & 67.5 & 67.3 & 58.3 \\
\bottomrule
\end{tabular}
}
\end{table}

\begin{table}[h]
\centering
\caption{\textbf{Performance of Qwen2.5-VL and other models on counting.}}
\label{tab:grounding_results}
\scalebox{0.92}{
\setlength{\tabcolsep}{3.0pt}
\begin{tabular}{@{}lcccccc@{}}
\toprule
\textbf{Datasets}       & \textbf{Gemini 1.5-Pro} & \textbf{GPT-4o} & \textbf{Claude-3.5 Sonnet}& \textbf{Molmo-72b}& \textbf{InternVL2.5-78B}&\textbf{Qwen2.5-VL-72B}  \\ 
\midrule
CountBench & 85.5   & 87.9 & 89.7 & 91.2 & 72.1 & 93.6  \\
\bottomrule
\end{tabular}
}
\end{table}

\subsubsection{Video Understanding and Grounding}

We assessed our models across a diverse range of video understanding and grounding tasks, utilizing benchmarks that include videos ranging from a few seconds to several hours in length.
Table~\ref{tab:video_results} demonstrates the performance comparison between Qwen2.5-VL models and top-tier proprietary models on the following video benchmarks: Video-MME~\citep{fu2024video}, Video-MMMU~\citep{hu2025video}, MMVU~\citep{zhao2025mmvu}, MVBench~\citep{li2024mvbench}, MMBench-Video~\citep{fang2024mmbench}, LongVideoBench~\citep{wu2024longvideobench}, EgoSchema~\citep{mangalam2023egoschema}, PerceptionTest~\citep{patraucean2024perception}, MLVU~\citep{zhou2024mlvu}, LVBench~\citep{wang2024lvbench}, TempCompass~\citep{liu2024tempcompass} and Charades-STA~\citep{gao2017tall}.
Notably, on LVBench and MLVU, which evaluate long-form video understanding capabilities through question-answering tasks, Qwen2.5-VL-72B achieves remarkable results, significantly outperforming strong competitors such as GPT-4o.

By utilizing the proposed synchronized MRoPE, Qwen2.5-VL enhances its capabilities in time-sensitive video understanding, featuring improved timestamp referencing, temporal grounding, dense captioning, and additional functionalities. On the Charades-STA dataset, which assesses the capability to accurately localize events or activities with precise timestamps, Qwen2.5-VL-72B achieves an impressive mIoU score of 50.9, thereby surpassing the performance of GPT-4o.
For all evaluated benchmarks, we capped the maximum number of frames analyzed per video at 768, with the total number of video tokens not exceeding 24,576.

\begin{table}[h]
\centering
\caption{\textbf{Performance of Qwen2.5-VL and other models on video benchmarks.}}
\label{tab:video_results}
\setlength{\tabcolsep}{3.0pt}
\begin{tabular}{@{}lccccc@{}}
\toprule
\textbf{Datasets}       & \textbf{Gemini 1.5-Pro} & \textbf{GPT-4o} & \textbf{Qwen2.5-VL-72B} & \textbf{Qwen2.5-VL-7B} & \textbf{Qwen2.5-VL-3B} \\ 
\midrule
\multicolumn{6}{c}{\textit{Video Understanding Tasks}} \\
\midrule
Video-MME$_{\text{w/o\ sub.}}$ & \textbf{75.0} & 71.9 & 73.3 & 65.1 & 61.5  \\ 
Video-MME$_{\text{w\ sub.}}$   & \textbf{81.3} & 77.2 & 79.1 & 71.6 & 67.6  \\ 
Video-MMMU                     & 53.9 & \textbf{61.2} & 60.2 & 47.4 & -     \\     
MMVU$_{\text{val}}$            & 65.4 & \textbf{67.4} & 62.9 & 50.1 & -     \\     
MVBench                        & 60.5 & 64.6 & \textbf{70.4} & 69.6 & 67.0  \\ 
MMBench-Video                  & 1.30 & 1.63 & \textbf{2.02} & 1.79 & 1.63  \\ 
LongVideoBench$_{\text{val}}$  & 64.0 & \textbf{66.7} & 60.7 & 56.0 & 54.2  \\ 
LVBench                        & 33.1 & 30.8 & \textbf{47.3} & 45.3 & 43.3  \\ 
EgoSchema$_{\text{test}}$      & 71.2 & 72.2 & \textbf{76.2} & 65.0 & 64.8  \\ 
PerceptionTest$_{\text{test}}$ & -    & -    & \textbf{73.2} & 70.5 & 66.9  \\ 
MLVU$_{\text{M-Avg}}$          & -    & 64.6 & \textbf{74.6} & 70.2 & 68.2  \\ 
TempCompass$_{\text{Avg}}$     & 67.1 & 73.8 & \textbf{74.8} & 71.7 & 64.4  \\ 
\midrule
\multicolumn{6}{c}{\textit{Video Grounding Tasks}} \\
\midrule
Charades-STA$_{\text{mIoU}}$   & -    & 35.7 & \textbf{50.9} & 43.6 & 38.8  \\ 
\bottomrule
\end{tabular}
\end{table}

\subsubsection{Agent}

Agent capabilities within multimodal models are crucial for enabling these models to effectively interact with real-world devices. We assess the agent capabilities of Qwen2.5-VL through various aspects. The UI elements grounding is evaluated by ScreenSpot~\citep{cheng2024seeclick} and ScreenSpot Pro~\citep{screenspotpro}. Offline evaluations are conducted on Android Control~\citep{li2024effects}, while online evaluations are performed on platforms including AndroidWorld~\citep{rawles2024androidworld}, MobileMiniWob++~\citep{rawles2024androidworld}, and OSWorld~\citep{xie2025osworld}. We compare the performance of Qwen2.5-VL-72B againsts other prominent models, such as GPT-4o~\citep{gpt4o}, Gemini 2.0~\citep{gemini2}, Claude~\citep{sonnet3_5_computer_use}, Aguvis-72B~\citep{xu2024aguvis}, and Qwen2-VL-72B~\citep{Qwen2-VL}. The results are demonstrated in \Cref{tab:agent_bench}.

\begin{table}[h]
\centering
\caption{\textbf{Performance of Qwen2.5-VL and other models on GUI Agent benchmarks.}}
\label{tab:agent_bench}
\resizebox{\textwidth}{!}{%
\setlength{\tabcolsep}{3.0pt}
\begin{tabular}{@{}lcccccc@{}}
\toprule
\textbf{Benchmarks} & \textbf{GPT-4o} & \textbf{Gemini 2.0} & \textbf{Claude} & \textbf{Aguvis-72B} & \textbf{Qwen2-VL-72B} & \textbf{Qwen2.5-VL-72B} \\ 
\midrule
ScreenSpot & 18.1 & 84.0 & 83.0 & \textbf{89.2} & - & 87.1 \\ 
ScreenSpot Pro & - & - & 17.1 & 23.6 & 1.6 & \textbf{43.6} \\  
Android Control High$_\text{EM}$ & 20.8 & 28.5 & 12.5 & 66.4 & 59.1 & \textbf{67.36} \\  
Android Control Low$_\text{EM}$ & 19.4 & 60.2 & 19.4 & 84.4 & 59.2 & \textbf{93.7} \\  
AndroidWorld$_\text{SR}$ & 34.5\% (SoM) & 26\% (SoM) & 27.9\% & 26.1\% & 6\% (SoM) & \textbf{35\%} \\  
MobileMiniWob++$_\text{SR}$ & 61\% & 42\% (SoM) & 61\% (SoM) & 66\% & 50\% (SoM) & \textbf{68\%} \\  
OSWorld & 5.03 & 4.70 & \textbf{14.90} & 10.26 & 2.42 & 8.83 \\  
\bottomrule
\end{tabular}
}
\end{table}

The performance of Qwen2.5-VL-72B demonstrates exceptional advancements across GUI grounding benchmarks. It achieves 87.1\% accuracy on ScreenSpot, competing strongly with Gemini 2.0 (84.0\%) and Claude (83.0\%), while notably setting a new standard on ScreenSpot Pro with 43.6\% accuracy - far surpassing both Aguvis-72B (23.6\%) and its foundation Qwen2-VL-72B (1.6\%). Leveraging these superior grounding capabilities, Qwen2.5-VL-72B significantly outperforms baselines across all offline evaluation benchmarks with a large gap. In online evaluation, some baselines have difficulty completing tasks due to limited grounding capabilities. Thus, we apply the Set-of-Mark (SoM) to the inputs of these models. The results show that Qwen2.5-VL-72B can outperform the baselines on AndroidWorld and MobileMiniWob++ and achieve comparable performance on OSWorld in online evaluation without auxiliary marks. This observation suggests that Qwen2.5-VL-72B is able to function as an agent in real and dynamic environments.
\section{Conclusion}
We present Qwen2.5-VL, a state-of-the-art vision-language model series that achieves significant advancements in multimodal understanding and interaction. With enhanced capabilities in visual recognition, object localization, document parsing, and long-video comprehension, Qwen2.5-VL excels in both static and dynamic tasks. Its native dynamic-resolution processing and absolute time encoding enable robust handling of diverse inputs, while Window Attention reduces computational overhead without sacrificing resolution fidelity. Qwen2.5-VL caters to a wide range of applications, from edge AI to high-performance computing. The flagship Qwen2.5-VL-72B matches or surpasses leading models like GPT-4o, and Claude 3.5 Sonnet, particularly in document and diagram understanding, while maintaining strong performance on pure text tasks. The smaller Qwen2.5-VL-7B and Qwen2.5-VL-3B variants outperform similarly sized competitors, offering efficiency and versatility. Qwen2.5-VL sets a new benchmark for vision-language models, demonstrating exceptional generalization and task execution across domains. Its innovations pave the way for more intelligent and interactive systems, bridging perception and real-world application. 

\section{Authors}
\textbf{Core Contributors:} Shuai Bai, Keqin Chen, Xuejing Liu,  Jialin Wang, Wenbin Ge, Sibo Song, Kai Dang, Peng Wang, Shijie Wang, Jun Tang, Humen Zhong, Yuanzhi Zhu, Mingkun Yang, Zhaohai Li, Jianqiang Wan, Pengfei Wang, Wei Ding, Zheren Fu, Yiheng Xu, Jiabo Ye, Xi Zhang, Tianbao Xie, Zesen Cheng, Hang Zhang, Zhibo Yang, Haiyang Xu, Junyang Lin

\textbf{Contributors\footnote{Alphabetical order.}:} An Yang, Binyuan Hui, Bowen Yu, Chen Cheng, Dayiheng Liu, Fan Hong, Fei Huang, Jiawei Liu, Jin Xu, Jianhong Tu, Jianyuan Zeng, Jie Zhang, Jinkai Wang, Jianwei Zhang, Jingren Zhou, Kexin Yang, Mei Li, Ming Yan, Na Ni, Rui Men, Songtao Jiang, Xiaodong Deng, Xiaoming Huang, Ximing Zhou, Xingzhang Ren, Yang Fan, Yichang Zhang, Yikai Zhu, Yuqiong Liu, Zhifang Guo

\bibliography{colm2024_conference}

\begin{thebibliography}{120}
\providecommand{\natexlab}[1]{#1}
\providecommand{\url}[1]{\texttt{#1}}
\expandafter\ifx\csname urlstyle\endcsname\relax
  \providecommand{\doi}[1]{doi: #1}\else
  \providecommand{\doi}{doi: \begingroup \urlstyle{rm}\Url}\fi

\bibitem[Agrawal et~al.(2024)Agrawal, Antoniak, Hanna, Bout, Chaplot, Chudnovsky, Costa, De~Monicault, Garg, Gervet, et~al.]{agrawal2024pixtral}
Pravesh Agrawal, Szymon Antoniak, Emma~Bou Hanna, Baptiste Bout, Devendra Chaplot, Jessica Chudnovsky, Diogo Costa, Baudouin De~Monicault, Saurabh Garg, Theophile Gervet, et~al.
\newblock Pixtral 12b.
\newblock \emph{arXiv preprint arXiv:2410.07073}, 2024.

\bibitem[Alayrac et~al.(2022)Alayrac, Donahue, Luc, Miech, Barr, Hasson, Lenc, Mensch, Millican, Reynolds, et~al.]{flamingo}
Jean-Baptiste Alayrac, Jeff Donahue, Pauline Luc, Antoine Miech, Iain Barr, Yana Hasson, Karel Lenc, Arthur Mensch, Katherine Millican, Malcolm Reynolds, et~al.
\newblock Flamingo: a visual language model for few-shot learning.
\newblock In \emph{NeurIPS}, 2022.

\bibitem[Anthropic(2024{\natexlab{a}})]{sonnet3_5}
Anthropic.
\newblock Claude 3.5 sonnet, 2024{\natexlab{a}}.
\newblock URL \url{https://www.anthropic.com/news/claude-3-5-sonnet}.

\bibitem[Anthropic(2024{\natexlab{b}})]{sonnet3_5_computer_use}
Anthropic.
\newblock Introducing computer use, a new claude 3.5 sonnet, and claude 3.5 haiku, 2024{\natexlab{b}}.
\newblock URL \url{https://www.anthropic.com/news/3-5-models-and-computer-use}.

\bibitem[Cassano et~al.(2023)Cassano, Gouwar, Nguyen, Nguyen, Phipps{-}Costin, Pinckney, Yee, Zi, Anderson, Feldman, Guha, Greenberg, and Jangda]{multiple}
Federico Cassano, John Gouwar, Daniel Nguyen, Sydney Nguyen, Luna Phipps{-}Costin, Donald Pinckney, Ming{-}Ho Yee, Yangtian Zi, Carolyn~Jane Anderson, Molly~Q. Feldman, Arjun Guha, Michael Greenberg, and Abhinav Jangda.
\newblock {MultiPL-E}: {A} scalable and polyglot approach to benchmarking neural code generation.
\newblock \emph{{IEEE} Trans. Software Eng.}, 49\penalty0 (7):\penalty0 3675--3691, 2023.

\bibitem[Chen et~al.(2024{\natexlab{a}})Chen, Chen, Zhang, Chen, Wu, Zhang, Chen, Li, Wan, and Wang]{chen2024allava}
Guiming~Hardy Chen, Shunian Chen, Ruifei Zhang, Junying Chen, Xiangbo Wu, Zhiyi Zhang, Zhihong Chen, Jianquan Li, Xiang Wan, and Benyou Wang.
\newblock Allava: Harnessing gpt4v-synthesized data for a lite vision-language model.
\newblock \emph{arXiv preprint arXiv:2402.11684}, 2024{\natexlab{a}}.

\bibitem[Chen et~al.(2024{\natexlab{b}})Chen, Liang, Siu, Wang, Wang, Wang, Ni, Zhu, Jiang, Lyu, et~al.]{chen2024mega}
Jiacheng Chen, Tianhao Liang, Sherman Siu, Zhengqing Wang, Kai Wang, Yubo Wang, Yuansheng Ni, Wang Zhu, Ziyan Jiang, Bohan Lyu, et~al.
\newblock Mega-bench: Scaling multimodal evaluation to over 500 real-world tasks.
\newblock \emph{arXiv preprint arXiv:2410.10563}, 2024{\natexlab{b}}.

\bibitem[Chen et~al.(2024{\natexlab{c}})Chen, Li, Dong, Zhang, Zang, Chen, Duan, Wang, Qiao, Lin, et~al.]{chen2024we}
Lin Chen, Jinsong Li, Xiaoyi Dong, Pan Zhang, Yuhang Zang, Zehui Chen, Haodong Duan, Jiaqi Wang, Yu~Qiao, Dahua Lin, et~al.
\newblock Are we on the right way for evaluating large vision-language models?
\newblock \emph{arXiv:2403.20330}, 2024{\natexlab{c}}.

\bibitem[Chen et~al.(2021)Chen, Tworek, Jun, Yuan, de~Oliveira~Pinto, Kaplan, Edwards, Burda, Joseph, Brockman, Ray, Puri, Krueger, Petrov, Khlaaf, Sastry, Mishkin, Chan, Gray, Ryder, Pavlov, Power, Kaiser, Bavarian, Winter, Tillet, Such, Cummings, Plappert, Chantzis, Barnes, Herbert{-}Voss, Guss, Nichol, Paino, Tezak, Tang, Babuschkin, Balaji, Jain, Saunders, Hesse, Carr, Leike, Achiam, Misra, Morikawa, Radford, Knight, Brundage, Murati, Mayer, Welinder, McGrew, Amodei, McCandlish, Sutskever, and Zaremba]{humaneval}
Mark Chen, Jerry Tworek, Heewoo Jun, Qiming Yuan, Henrique~Pond{\'{e}} de~Oliveira~Pinto, Jared Kaplan, Harrison Edwards, Yuri Burda, Nicholas Joseph, Greg Brockman, Alex Ray, Raul Puri, Gretchen Krueger, Michael Petrov, Heidy Khlaaf, Girish Sastry, Pamela Mishkin, Brooke Chan, Scott Gray, Nick Ryder, Mikhail Pavlov, Alethea Power, Lukasz Kaiser, Mohammad Bavarian, Clemens Winter, Philippe Tillet, Felipe~Petroski Such, Dave Cummings, Matthias Plappert, Fotios Chantzis, Elizabeth Barnes, Ariel Herbert{-}Voss, William~Hebgen Guss, Alex Nichol, Alex Paino, Nikolas Tezak, Jie Tang, Igor Babuschkin, Suchir Balaji, Shantanu Jain, William Saunders, Christopher Hesse, Andrew~N. Carr, Jan Leike, Joshua Achiam, Vedant Misra, Evan Morikawa, Alec Radford, Matthew Knight, Miles Brundage, Mira Murati, Katie Mayer, Peter Welinder, Bob McGrew, Dario Amodei, Sam McCandlish, Ilya Sutskever, and Wojciech Zaremba.
\newblock Evaluating large language models trained on code.
\newblock \emph{CoRR}, abs/2107.03374, 2021.

\bibitem[Chen et~al.(2023)Chen, Wu, Wang, Su, Chen, Xing, Zhong, Zhang, Zhu, Lu, Li, Luo, Lu, Qiao, and Dai]{internvl}
Zhe Chen, Jiannan Wu, Wenhai Wang, Weijie Su, Guo Chen, Sen Xing, Muyan Zhong, Qinglong Zhang, Xizhou Zhu, Lewei Lu, Bin Li, Ping Luo, Tong Lu, Yu~Qiao, and Jifeng Dai.
\newblock Internvl: Scaling up vision foundation models and aligning for generic visual-linguistic tasks.
\newblock \emph{arXiv preprint arXiv:2312.14238}, 2023.

\bibitem[Chen et~al.(2024{\natexlab{d}})Chen, Wang, Cao, Liu, Gao, Cui, Zhu, Ye, Tian, Liu, et~al.]{chen2024expanding}
Zhe Chen, Weiyun Wang, Yue Cao, Yangzhou Liu, Zhangwei Gao, Erfei Cui, Jinguo Zhu, Shenglong Ye, Hao Tian, Zhaoyang Liu, et~al.
\newblock Expanding performance boundaries of open-source multimodal models with model, data, and test-time scaling.
\newblock \emph{arXiv preprint arXiv:2412.05271}, 2024{\natexlab{d}}.

\bibitem[Cheng et~al.(2024)Cheng, Sun, Chu, Xu, Li, Zhang, and Wu]{cheng2024seeclick}
Kanzhi Cheng, Qiushi Sun, Yougang Chu, Fangzhi Xu, Yantao Li, Jianbing Zhang, and Zhiyong Wu.
\newblock Seeclick: Harnessing gui grounding for advanced visual gui agents.
\newblock \emph{arXiv preprint arXiv:2401.10935}, 2024.

\bibitem[Cobbe et~al.(2021)Cobbe, Kosaraju, Bavarian, Chen, Jun, Kaiser, Plappert, Tworek, Hilton, Nakano, Hesse, and Schulman]{gsm8k}
Karl Cobbe, Vineet Kosaraju, Mohammad Bavarian, Mark Chen, Heewoo Jun, Lukasz Kaiser, Matthias Plappert, Jerry Tworek, Jacob Hilton, Reiichiro Nakano, Christopher Hesse, and John Schulman.
\newblock Training verifiers to solve math word problems.
\newblock \emph{CoRR}, abs/2110.14168, 2021.

\bibitem[Dauphin et~al.(2017)Dauphin, Fan, Auli, and Grangier]{glu}
Yann~N. Dauphin, Angela Fan, Michael Auli, and David Grangier.
\newblock Language modeling with gated convolutional networks.
\newblock In \emph{{ICML}}, volume~70 of \emph{Proceedings of Machine Learning Research}, pp.\  933--941. {PMLR}, 2017.

\bibitem[Deepmind(2024)]{gemini2}
Google Deepmind.
\newblock Introducing gemini 2.0: our new ai model for the agentic era, 2024.
\newblock URL \url{https://blog.google/technology/google-deepmind/google-gemini-ai-update-december-2024/}.

\bibitem[DeepSeek{-}AI et~al.(2024)DeepSeek{-}AI, Liu, Feng, Xue, Wang, Wu, Lu, Zhao, Deng, Zhang, Ruan, Dai, Guo, Yang, Chen, Ji, Li, Lin, Dai, Luo, Hao, Chen, Li, Zhang, Bao, Xu, Wang, Zhang, Ding, Xin, Gao, Li, Qu, Cai, Liang, Guo, Ni, Li, Wang, Chen, Chen, Yuan, Qiu, Li, Song, Dong, Hu, Gao, Guan, Huang, Yu, Wang, Zhang, Xu, Xia, Zhao, Wang, Zhang, Li, Wang, Zhang, Zhang, Tang, Li, Tian, Huang, Wang, Zhang, Wang, Zhu, Chen, Du, Chen, Jin, Ge, Zhang, Pan, Wang, Xu, Zhang, Chen, Li, Lu, Zhou, Chen, Wu, Ye, Ye, Ma, Wang, Zhou, Yu, Zhou, Pan, Wang, Yun, Pei, Sun, Xiao, and Zeng]{DBLP:journals/corr/abs-2412-19437}
DeepSeek{-}AI, Aixin Liu, Bei Feng, Bing Xue, Bingxuan Wang, Bochao Wu, Chengda Lu, Chenggang Zhao, Chengqi Deng, Chenyu Zhang, Chong Ruan, Damai Dai, Daya Guo, Dejian Yang, Deli Chen, Dongjie Ji, Erhang Li, Fangyun Lin, Fucong Dai, Fuli Luo, Guangbo Hao, Guanting Chen, Guowei Li, H.~Zhang, Han Bao, Hanwei Xu, Haocheng Wang, Haowei Zhang, Honghui Ding, Huajian Xin, Huazuo Gao, Hui Li, Hui Qu, J.~L. Cai, Jian Liang, Jianzhong Guo, Jiaqi Ni, Jiashi Li, Jiawei Wang, Jin Chen, Jingchang Chen, Jingyang Yuan, Junjie Qiu, Junlong Li, Junxiao Song, Kai Dong, Kai Hu, Kaige Gao, Kang Guan, Kexin Huang, Kuai Yu, Lean Wang, Lecong Zhang, Lei Xu, Leyi Xia, Liang Zhao, Litong Wang, Liyue Zhang, Meng Li, Miaojun Wang, Mingchuan Zhang, Minghua Zhang, Minghui Tang, Mingming Li, Ning Tian, Panpan Huang, Peiyi Wang, Peng Zhang, Qiancheng Wang, Qihao Zhu, Qinyu Chen, Qiushi Du, R.~J. Chen, R.~L. Jin, Ruiqi Ge, Ruisong Zhang, Ruizhe Pan, Runji Wang, Runxin Xu, Ruoyu Zhang, Ruyi Chen, S.~S. Li, Shanghao Lu, Shangyan Zhou,
  Shanhuang Chen, Shaoqing Wu, Shengfeng Ye, Shengfeng Ye, Shirong Ma, Shiyu Wang, Shuang Zhou, Shuiping Yu, Shunfeng Zhou, Shuting Pan, T.~Wang, Tao Yun, Tian Pei, Tianyu Sun, W.~L. Xiao, and Wangding Zeng.
\newblock Deepseek-v3 technical report.
\newblock \emph{CoRR}, abs/2412.19437, 2024.
\newblock \doi{10.48550/ARXIV.2412.19437}.
\newblock URL \url{https://doi.org/10.48550/arXiv.2412.19437}.

\bibitem[Deitke et~al.(2024)Deitke, Clark, Lee, Tripathi, Yang, Park, Salehi, Muennighoff, Lo, Soldaini, et~al.]{deitke2024molmo}
Matt Deitke, Christopher Clark, Sangho Lee, Rohun Tripathi, Yue Yang, Jae~Sung Park, Mohammadreza Salehi, Niklas Muennighoff, Kyle Lo, Luca Soldaini, et~al.
\newblock Molmo and pixmo: Open weights and open data for state-of-the-art multimodal models.
\newblock \emph{arXiv preprint arXiv:2409.17146}, 2024.

\bibitem[Fang et~al.(2024)Fang, Mao, Duan, Zhao, Li, Lin, and Chen]{fang2024mmbench}
Xinyu Fang, Kangrui Mao, Haodong Duan, Xiangyu Zhao, Yining Li, Dahua Lin, and Kai Chen.
\newblock Mmbench-video: A long-form multi-shot benchmark for holistic video understanding.
\newblock \emph{arXiv preprint arXiv:2406.14515}, 2024.

\bibitem[Fu et~al.(2023)Fu, Chen, Shen, Qin, Zhang, Lin, Qiu, Lin, Yang, Zheng, et~al.]{fu2023mme}
Chaoyou Fu, Peixian Chen, Yunhang Shen, Yulei Qin, Mengdan Zhang, Xu~Lin, Zhenyu Qiu, Wei Lin, Jinrui Yang, Xiawu Zheng, et~al.
\newblock Mme: A comprehensive evaluation benchmark for multimodal large language models.
\newblock \emph{arXiv:2306.13394}, 2023.

\bibitem[Fu et~al.(2024{\natexlab{a}})Fu, Dai, Luo, Li, Ren, Zhang, Wang, Zhou, Shen, Zhang, et~al.]{fu2024video}
Chaoyou Fu, Yuhan Dai, Yondong Luo, Lei Li, Shuhuai Ren, Renrui Zhang, Zihan Wang, Chenyu Zhou, Yunhang Shen, Mengdan Zhang, et~al.
\newblock Video-mme: The first-ever comprehensive evaluation benchmark of multi-modal llms in video analysis.
\newblock \emph{arXiv:2405.21075}, 2024{\natexlab{a}}.

\bibitem[Fu et~al.(2024{\natexlab{b}})Fu, Yang, Kuang, Song, Li, Zhu, Luo, Wang, Lu, Huang, Li, Tang, Shan, Lin, Liu, Wu, Feng, Liu, Huang, Tang, Chen, Jin, Liu, and Bai]{fu2024ocrbenchv2improvedbenchmark}
Ling Fu, Biao Yang, Zhebin Kuang, Jiajun Song, Yuzhe Li, Linghao Zhu, Qidi Luo, Xinyu Wang, Hao Lu, Mingxin Huang, Zhang Li, Guozhi Tang, Bin Shan, Chunhui Lin, Qi~Liu, Binghong Wu, Hao Feng, Hao Liu, Can Huang, Jingqun Tang, Wei Chen, Lianwen Jin, Yuliang Liu, and Xiang Bai.
\newblock Ocrbench v2: An improved benchmark for evaluating large multimodal models on visual text localization and reasoning, 2024{\natexlab{b}}.
\newblock URL \url{https://arxiv.org/abs/2501.00321}.

\bibitem[Fu et~al.(2024{\natexlab{c}})Fu, Hu, Li, Feng, Wang, Lin, Roth, Smith, Ma, and Krishna]{fu2024blink}
Xingyu Fu, Yushi Hu, Bangzheng Li, Yu~Feng, Haoyu Wang, Xudong Lin, Dan Roth, Noah~A Smith, Wei-Chiu Ma, and Ranjay Krishna.
\newblock Blink: Multimodal large language models can see but not perceive.
\newblock In \emph{European Conference on Computer Vision}, pp.\  148--166. Springer, 2024{\natexlab{c}}.

\bibitem[Gadre et~al.(2023)Gadre, Ilharco, Fang, Hayase, Smyrnis, Nguyen, Marten, Wortsman, Ghosh, Zhang, et~al.]{datacomp}
Samir~Yitzhak Gadre, Gabriel Ilharco, Alex Fang, Jonathan Hayase, Georgios Smyrnis, Thao Nguyen, Ryan Marten, Mitchell Wortsman, Dhruba Ghosh, Jieyu Zhang, et~al.
\newblock Datacomp: In search of the next generation of multimodal datasets.
\newblock \emph{arXiv:2304.14108}, 2023.

\bibitem[Gao et~al.(2017)Gao, Sun, Yang, and Nevatia]{gao2017tall}
Jiyang Gao, Chen Sun, Zhenheng Yang, and Ram Nevatia.
\newblock Tall: Temporal activity localization via language query.
\newblock In \emph{Proceedings of the IEEE international conference on computer vision}, pp.\  5267--5275, 2017.

\bibitem[Gema et~al.(2024)Gema, Leang, Hong, Devoto, Mancino, Saxena, He, Zhao, Du, Madani, et~al.]{mmluredux}
Aryo~Pradipta Gema, Joshua Ong~Jun Leang, Giwon Hong, Alessio Devoto, Alberto Carlo~Maria Mancino, Rohit Saxena, Xuanli He, Yu~Zhao, Xiaotang Du, Mohammad Reza~Ghasemi Madani, et~al.
\newblock Are we done with mmlu?
\newblock \emph{CoRR}, abs/2406.04127, 2024.

\bibitem[Ghiasi et~al.(2021)Ghiasi, Cui, Srinivas, Qian, Lin, Cubuk, Le, and Zoph]{ghiasi2021simple}
Golnaz Ghiasi, Yin Cui, Aravind Srinivas, Rui Qian, Tsung-Yi Lin, Ekin~D Cubuk, Quoc~V Le, and Barret Zoph.
\newblock Simple copy-paste is a strong data augmentation method for instance segmentation.
\newblock In \emph{Proceedings of the IEEE/CVF conference on computer vision and pattern recognition}, pp.\  2918--2928, 2021.

\bibitem[Guan et~al.(2023)Guan, Liu, Wu, Xian, Li, Liu, Wang, Chen, Huang, Yacoob, Manocha, and Zhou]{guan2023hallusionbench}
Tianrui Guan, Fuxiao Liu, Xiyang Wu, Ruiqi Xian, Zongxia Li, Xiaoyu Liu, Xijun Wang, Lichang Chen, Furong Huang, Yaser Yacoob, Dinesh Manocha, and Tianyi Zhou.
\newblock Hallusionbench: An advanced diagnostic suite for entangled language hallucination \& visual illusion in large vision-language models.
\newblock \emph{arXiv:2310.14566}, 2023.

\bibitem[Guo et~al.(2024)Guo, Zheng, Bai, Li, Wang, Zhu, Li, Neubig, Chen, and Yue]{guo2024mammoth}
Jarvis Guo, Tuney Zheng, Yuelin Bai, Bo~Li, Yubo Wang, King Zhu, Yizhi Li, Graham Neubig, Wenhu Chen, and Xiang Yue.
\newblock Mammoth-vl: Eliciting multimodal reasoning with instruction tuning at scale.
\newblock \emph{arXiv preprint arXiv:2412.05237}, 2024.

\bibitem[Hendrycks et~al.(2021)Hendrycks, Burns, Kadavath, Arora, Basart, Tang, Song, and Steinhardt]{math}
Dan Hendrycks, Collin Burns, Saurav Kadavath, Akul Arora, Steven Basart, Eric Tang, Dawn Song, and Jacob Steinhardt.
\newblock Measuring mathematical problem solving with the {MATH} dataset.
\newblock In \emph{NeurIPS Datasets and Benchmarks}, 2021.

\bibitem[Hu et~al.(2025)Hu, Wu, Pu, Xiao, Zhang, Yue, Li, and Liu]{hu2025video}
Kairui Hu, Penghao Wu, Fanyi Pu, Wang Xiao, Yuanhan Zhang, Xiang Yue, Bo~Li, and Ziwei Liu.
\newblock Video-mmmu: Evaluating knowledge acquisition from multi-discipline professional videos.
\newblock \emph{arXiv preprint arXiv:2501.13826}, 2025.

\bibitem[Kazemzadeh et~al.(2014)Kazemzadeh, Ordonez, Matten, and Berg]{refcoco}
Sahar Kazemzadeh, Vicente Ordonez, Mark Matten, and Tamara Berg.
\newblock Referitgame: Referring to objects in photographs of natural scenes.
\newblock In \emph{EMNLP}, 2014.

\bibitem[Kembhavi et~al.(2016)Kembhavi, Salvato, Kolve, Seo, Hajishirzi, and Farhadi]{kembhavi2016diagram}
Aniruddha Kembhavi, Mike Salvato, Eric Kolve, Minjoon Seo, Hannaneh Hajishirzi, and Ali Farhadi.
\newblock A diagram is worth a dozen images.
\newblock In \emph{ECCV}, 2016.

\bibitem[Kirillov et~al.(2023)Kirillov, Mintun, Ravi, Mao, Rolland, Gustafson, Xiao, Whitehead, Berg, Lo, et~al.]{kirillov2023segment}
Alexander Kirillov, Eric Mintun, Nikhila Ravi, Hanzi Mao, Chloe Rolland, Laura Gustafson, Tete Xiao, Spencer Whitehead, Alexander~C Berg, Wan-Yen Lo, et~al.
\newblock Segment anything.
\newblock In \emph{ICCV}, 2023.

\bibitem[Lee et~al.(2024)Lee, Park, Won~Kim, and Man~Ro]{lee2024moai}
Byung-Kwan Lee, Beomchan Park, Chae Won~Kim, and Yong Man~Ro.
\newblock Moai: Mixture of all intelligence for large language and vision models.
\newblock In \emph{European Conference on Computer Vision}, pp.\  273--302. Springer, 2024.

\bibitem[Li et~al.(2023{\natexlab{a}})Li, Zhang, Yang, Zhang, Pu, and Liu]{Otterhd}
Bo~Li, Peiyuan Zhang, Jingkang Yang, Yuanhan Zhang, Fanyi Pu, and Ziwei Liu.
\newblock Otterhd: A high-resolution multi-modality model.
\newblock \emph{arXiv:2311.04219}, 2023{\natexlab{a}}.

\bibitem[Li et~al.(2024{\natexlab{a}})Li, Zhang, Guo, Zhang, Li, Zhang, Zhang, Zhang, Li, Liu, et~al.]{li2024llava}
Bo~Li, Yuanhan Zhang, Dong Guo, Renrui Zhang, Feng Li, Hao Zhang, Kaichen Zhang, Peiyuan Zhang, Yanwei Li, Ziwei Liu, et~al.
\newblock Llava-onevision: Easy visual task transfer.
\newblock \emph{arXiv preprint arXiv:2408.03326}, 2024{\natexlab{a}}.

\bibitem[Li et~al.(2024{\natexlab{b}})Li, Ge, Chen, Ge, Zhang, and Shan]{li2024seed2plus}
Bohao Li, Yuying Ge, Yi~Chen, Yixiao Ge, Ruimao Zhang, and Ying Shan.
\newblock Seed-bench-2-plus: Benchmarking multimodal large language models with text-rich visual comprehension.
\newblock \emph{arXiv preprint arXiv:2404.16790}, 2024{\natexlab{b}}.

\bibitem[Li et~al.(2024{\natexlab{c}})Li, Liu, Wu, Wang, Shen, Qu, Niu, Wang, Chen, and Li]{li2024aria}
Dongxu Li, Yudong Liu, Haoning Wu, Yue Wang, Zhiqi Shen, Bowen Qu, Xinyao Niu, Guoyin Wang, Bei Chen, and Junnan Li.
\newblock Aria: An open multimodal native mixture-of-experts model.
\newblock \emph{arXiv preprint arXiv:2410.05993}, 2024{\natexlab{c}}.

\bibitem[Li et~al.(2022{\natexlab{a}})Li, Li, Xiong, and Hoi]{blip}
Junnan Li, Dongxu Li, Caiming Xiong, and Steven C.~H. Hoi.
\newblock Blip: Bootstrapping language-image pre-training for unified vision-language understanding and generation.
\newblock In \emph{ICML}, 2022{\natexlab{a}}.

\bibitem[Li et~al.(2023{\natexlab{b}})Li, Li, Savarese, and Hoi]{blip2}
Junnan Li, Dongxu Li, Silvio Savarese, and Steven Hoi.
\newblock Blip-2: Bootstrapping language-image pre-training with frozen image encoders and large language models.
\newblock \emph{arXiv:2301.12597}, 2023{\natexlab{b}}.

\bibitem[Li et~al.(2025{\natexlab{a}})Li, Meng, Lin, Luo, Tian, Ma, Huang, and Chua]{screenspotpro}
Kaixin Li, Ziyang Meng, Hongzhan Lin, Ziyang Luo, Yuchen Tian, Jing Ma, Zhiyong Huang, and Tat-Seng Chua.
\newblock Screenspot-pro: Gui grounding for professional high-resolution computer use, 2025{\natexlab{a}}.
\newblock URL \url{https://likaixin2000.github.io/papers/ScreenSpot_Pro.pdf}.
\newblock Preprint.

\bibitem[Li et~al.(2024{\natexlab{d}})Li, Wang, He, Li, Wang, Liu, Wang, Xu, Chen, Luo, et~al.]{li2024mvbench}
Kunchang Li, Yali Wang, Yinan He, Yizhuo Li, Yi~Wang, Yi~Liu, Zun Wang, Jilan Xu, Guo Chen, Ping Luo, et~al.
\newblock Mvbench: A comprehensive multi-modal video understanding benchmark.
\newblock In \emph{CVPR}, 2024{\natexlab{d}}.

\bibitem[Li et~al.(2022{\natexlab{b}})Li, Zhang, Zhang, Yang, Li, Zhong, Wang, Yuan, Zhang, Hwang, et~al.]{li2022grounded}
Liunian~Harold Li, Pengchuan Zhang, Haotian Zhang, Jianwei Yang, Chunyuan Li, Yiwu Zhong, Lijuan Wang, Lu~Yuan, Lei Zhang, Jenq-Neng Hwang, et~al.
\newblock Grounded language-image pre-training.
\newblock In \emph{Proceedings of the IEEE/CVF Conference on Computer Vision and Pattern Recognition}, pp.\  10965--10975, 2022{\natexlab{b}}.

\bibitem[Li et~al.(2024{\natexlab{e}})Li, Chen, Wang, Wang, Ye, Jin, Chen, He, Gao, Cui, et~al.]{li2024omnicorpus}
Qingyun Li, Zhe Chen, Weiyun Wang, Wenhai Wang, Shenglong Ye, Zhenjiang Jin, Guanzhou Chen, Yinan He, Zhangwei Gao, Erfei Cui, et~al.
\newblock Omnicorpus: An unified multimodal corpus of 10 billion-level images interleaved with text.
\newblock \emph{arXiv preprint arXiv:2406.08418}, 2024{\natexlab{e}}.

\bibitem[Li et~al.(2024{\natexlab{f}})Li, Bishop, Li, Rawles, Campbell-Ajala, Tyamagundlu, and Riva]{li2024effects}
Wei Li, William Bishop, Alice Li, Chris Rawles, Folawiyo Campbell-Ajala, Divya Tyamagundlu, and Oriana Riva.
\newblock On the effects of data scale on computer control agents.
\newblock \emph{arXiv preprint arXiv:2406.03679}, 2024{\natexlab{f}}.

\bibitem[Li et~al.(2024{\natexlab{g}})Li, Sun, Lin, Li, Dong, Zhang, Ding, Song, Cheng, Huo, et~al.]{li2024baichuan}
Yadong Li, Haoze Sun, Mingan Lin, Tianpeng Li, Guosheng Dong, Tao Zhang, Bowen Ding, Wei Song, Zhenglin Cheng, Yuqi Huo, et~al.
\newblock Baichuan-omni technical report.
\newblock \emph{arXiv preprint arXiv:2410.08565}, 3\penalty0 (7), 2024{\natexlab{g}}.

\bibitem[Li et~al.(2025{\natexlab{b}})Li, Liu, Zhang, Chen, Li, Li, Liu, Ming, Dong, Pan, et~al.]{li2025baichuan}
Yadong Li, Jun Liu, Tao Zhang, Song Chen, Tianpeng Li, Zehuan Li, Lijun Liu, Lingfeng Ming, Guosheng Dong, Da~Pan, et~al.
\newblock Baichuan-omni-1.5 technical report.
\newblock \emph{arXiv preprint arXiv:2501.15368}, 2025{\natexlab{b}}.

\bibitem[Li et~al.(2024{\natexlab{h}})Li, Jiang, Hu, Wang, Zhong, Luo, Ma, and Zhang]{li2024uni}
Yunxin Li, Shenyuan Jiang, Baotian Hu, Longyue Wang, Wanqi Zhong, Wenhan Luo, Lin Ma, and Min Zhang.
\newblock Uni-moe: Scaling unified multimodal llms with mixture of experts.
\newblock \emph{arXiv preprint arXiv:2405.11273}, 2024{\natexlab{h}}.

\bibitem[Li et~al.(2023{\natexlab{c}})Li, Yang, Liu, Ma, Zhang, Yang, Sun, Liu, and Bai]{monkey}
Zhang Li, Biao Yang, Qiang Liu, Zhiyin Ma, Shuo Zhang, Jingxu Yang, Yabo Sun, Yuliang Liu, and Xiang Bai.
\newblock Monkey: Image resolution and text label are important things for large multi-modal models.
\newblock \emph{arXiv:2311.06607}, 2023{\natexlab{c}}.

\bibitem[Liang et~al.(2025)Liang, Li, Chen, Chen, Zheng, Lai, Li, and Xue]{liang2025global}
Yuxuan Liang, Xu~Li, Xiaolei Chen, Haotian Chen, Yi~Zheng, Chenghang Lai, Bin Li, and Xiangyang Xue.
\newblock Global semantic-guided sub-image feature weight allocation in high-resolution large vision-language models.
\newblock \emph{arXiv preprint arXiv:2501.14276}, 2025.

\bibitem[Lin et~al.(2024)Lin, Yin, Ping, Molchanov, Shoeybi, and Han]{lin2024vila}
Ji~Lin, Hongxu Yin, Wei Ping, Pavlo Molchanov, Mohammad Shoeybi, and Song Han.
\newblock Vila: On pre-training for visual language models.
\newblock In \emph{Proceedings of the IEEE/CVF Conference on Computer Vision and Pattern Recognition}, pp.\  26689--26699, 2024.

\bibitem[Liu et~al.(2023{\natexlab{a}})Liu, Li, Li, and Lee]{llava1.5}
Haotian Liu, Chunyuan Li, Yuheng Li, and Yong~Jae Lee.
\newblock Improved baselines with visual instruction tuning.
\newblock \emph{arXiv:2310.03744}, 2023{\natexlab{a}}.

\bibitem[Liu et~al.(2023{\natexlab{b}})Liu, Li, Wu, and Lee]{llava}
Haotian Liu, Chunyuan Li, Qingyang Wu, and Yong~Jae Lee.
\newblock Visual instruction tuning.
\newblock \emph{arXiv:2304.08485}, 2023{\natexlab{b}}.

\bibitem[Liu et~al.(2023{\natexlab{c}})Liu, Zeng, Ren, Li, Zhang, Yang, yue Li, Yang, Su, Zhu, and Zhang]{grounding_dino}
Shilong Liu, Zhaoyang Zeng, Tianhe Ren, Feng Li, Hao Zhang, Jie Yang, Chun yue Li, Jianwei Yang, Hang Su, Jun-Juan Zhu, and Lei Zhang.
\newblock Grounding dino: Marrying dino with grounded pre-training for open-set object detection.
\newblock \emph{arXiv:2303.05499}, 2023{\natexlab{c}}.

\bibitem[Liu et~al.(2024{\natexlab{a}})Liu, Cao, Gao, Wang, Chen, Wang, Tian, Lu, Zhu, Lu, et~al.]{liu2024mminstruct}
Yangzhou Liu, Yue Cao, Zhangwei Gao, Weiyun Wang, Zhe Chen, Wenhai Wang, Hao Tian, Lewei Lu, Xizhou Zhu, Tong Lu, et~al.
\newblock Mminstruct: A high-quality multi-modal instruction tuning dataset with extensive diversity.
\newblock \emph{Science China Information Sciences}, 67\penalty0 (12):\penalty0 1--16, 2024{\natexlab{a}}.

\bibitem[Liu et~al.(2023{\natexlab{d}})Liu, Duan, Yuanhan~Zhang, Zhang, Zhao, Yuan, Wang, He, Liu, Chen, and Lin]{MMBench}
Yuan Liu, Haodong Duan, Bo~Li Yuanhan~Zhang, Songyang Zhang, Wangbo Zhao, Yike Yuan, Jiaqi Wang, Conghui He, Ziwei Liu, Kai Chen, and Dahua Lin.
\newblock Mmbench: Is your multi-modal model an all-around player?
\newblock \emph{arXiv:2307.06281}, 2023{\natexlab{d}}.

\bibitem[Liu et~al.(2024{\natexlab{b}})Liu, Zhao, Zhuang, Tian, Zhou, and Zhou]{liu2024points}
Yuan Liu, Zhongyin Zhao, Ziyuan Zhuang, Le~Tian, Xiao Zhou, and Jie Zhou.
\newblock Points: Improving your vision-language model with affordable strategies.
\newblock \emph{arXiv preprint arXiv:2409.04828}, 2024{\natexlab{b}}.

\bibitem[Liu et~al.(2024{\natexlab{c}})Liu, Li, Liu, Wang, Ren, Li, Chen, Sun, and Hou]{liu2024tempcompass}
Yuanxin Liu, Shicheng Li, Yi~Liu, Yuxiang Wang, Shuhuai Ren, Lei Li, Sishuo Chen, Xu~Sun, and Lu~Hou.
\newblock Tempcompass: Do video llms really understand videos?
\newblock \emph{arXiv preprint arXiv:2403.00476}, 2024{\natexlab{c}}.

\bibitem[Liu et~al.(2023{\natexlab{e}})Liu, Li, Huang, Yang, Yu, Li, Yin, lin Liu, Jin, and Bai]{liu2024ocrbenchhiddenmysteryocr}
Yuliang Liu, Zhang Li, Mingxin Huang, Biao Yang, Wenwen Yu, Chunyuan Li, Xucheng Yin, Cheng lin Liu, Lianwen Jin, and Xiang Bai.
\newblock Ocrbench: On the hidden mystery of ocr in large multimodal models.
\newblock \emph{arXiv:2305.07895}, 2023{\natexlab{e}}.

\bibitem[Lu et~al.(2024)Lu, Bansal, Xia, Liu, Li, Hajishirzi, Cheng, Chang, Galley, and Gao]{mathvista}
Pan Lu, Hritik Bansal, Tony Xia, Jiacheng Liu, Chunyuan Li, Hannaneh Hajishirzi, Hao Cheng, Kai{-}Wei Chang, Michel Galley, and Jianfeng Gao.
\newblock Mathvista: Evaluating mathematical reasoning of foundation models in visual contexts.
\newblock In \emph{ICLR}, 2024.

\bibitem[Mangalam et~al.(2023)Mangalam, Akshulakov, and Malik]{mangalam2023egoschema}
Karttikeya Mangalam, Raiymbek Akshulakov, and Jitendra Malik.
\newblock Egoschema: A diagnostic benchmark for very long-form video language understanding.
\newblock In \emph{NeurIPS}, 2023.

\bibitem[Mao et~al.(2016)Mao, Huang, Toshev, Camburu, Yuille, and Murphy]{refcocog}
Junhua Mao, Jonathan Huang, Alexander Toshev, Oana Camburu, Alan~L Yuille, and Kevin Murphy.
\newblock Generation and comprehension of unambiguous object descriptions.
\newblock In \emph{CVPR}, 2016.

\bibitem[Masry et~al.(2022)Masry, Long, Tan, Joty, and Hoque]{masry2022chartqa}
Ahmed Masry, Do~Xuan Long, Jia~Qing Tan, Shafiq Joty, and Enamul Hoque.
\newblock Chartqa: A benchmark for question answering about charts with visual and logical reasoning.
\newblock \emph{arXiv:2203.10244}, 2022.

\bibitem[Mathew et~al.(2021{\natexlab{a}})Mathew, Bagal, Tito, Karatzas, Valveny, and Jawahar]{Mathew2021InfographicVQA}
Minesh Mathew, Viraj Bagal, Rub{\`e}n~P{\'e}rez Tito, Dimosthenis Karatzas, Ernest Valveny, and C.V. Jawahar.
\newblock Infographicvqa.
\newblock \emph{2022 IEEE/CVF Winter Conference on Applications of Computer Vision (WACV)}, pp.\  2582--2591, 2021{\natexlab{a}}.

\bibitem[Mathew et~al.(2021{\natexlab{b}})Mathew, Karatzas, and Jawahar]{docvqa}
Minesh Mathew, Dimosthenis Karatzas, and CV~Jawahar.
\newblock Docvqa: A dataset for vqa on document images.
\newblock In \emph{WACV}, 2021{\natexlab{b}}.

\bibitem[MiniMax et~al.(2025)MiniMax, Li, Gong, Yang, Shan, Liu, Zhu, Zhang, Guo, Chen, Li, Jiao, Li, Zhang, Sun, Dong, Zhu, Zhuang, Song, Zhu, Han, Li, Xie, Xu, Yan, Zhang, Xiao, Kang, Han, Wang, Yu, Feng, Zheng, Chai, Xing, Ju, Chi, Zhang, Huang, Niu, Li, Zhao, Yang, Xu, Wang, Wang, Li, Leng, Shi, Yu, Li, Zhu, Huang, Liang, Sun, Sun, Cheng, Li, Song, Su, Han, Zhang, Hou, Min, Zou, Shen, Gong, Zhu, Zhou, Zhong, Hu, Fan, Yu, Yang, Li, Huang, Li, Huang, Xu, Mao, Li, Li, Tao, Ying, Cong, Qin, Fan, Yu, Jiang, and Wu]{minimax2025minimax01scalingfoundationmodels}
MiniMax, Aonian Li, Bangwei Gong, Bo~Yang, Boji Shan, Chang Liu, Cheng Zhu, Chunhao Zhang, Congchao Guo, Da~Chen, Dong Li, Enwei Jiao, Gengxin Li, Guojun Zhang, Haohai Sun, Houze Dong, Jiadai Zhu, Jiaqi Zhuang, Jiayuan Song, Jin Zhu, Jingtao Han, Jingyang Li, Junbin Xie, Junhao Xu, Junjie Yan, Kaishun Zhang, Kecheng Xiao, Kexi Kang, Le~Han, Leyang Wang, Lianfei Yu, Liheng Feng, Lin Zheng, Linbo Chai, Long Xing, Meizhi Ju, Mingyuan Chi, Mozhi Zhang, Peikai Huang, Pengcheng Niu, Pengfei Li, Pengyu Zhao, Qi~Yang, Qidi Xu, Qiexiang Wang, Qin Wang, Qiuhui Li, Ruitao Leng, Shengmin Shi, Shuqi Yu, Sichen Li, Songquan Zhu, Tao Huang, Tianrun Liang, Weigao Sun, Weixuan Sun, Weiyu Cheng, Wenkai Li, Xiangjun Song, Xiao Su, Xiaodong Han, Xinjie Zhang, Xinzhu Hou, Xu~Min, Xun Zou, Xuyang Shen, Yan Gong, Yingjie Zhu, Yipeng Zhou, Yiran Zhong, Yongyi Hu, Yuanxiang Fan, Yue Yu, Yufeng Yang, Yuhao Li, Yunan Huang, Yunji Li, Yunpeng Huang, Yunzhi Xu, Yuxin Mao, Zehan Li, Zekang Li, Zewei Tao, Zewen Ying, Zhaoyang Cong, Zhen
  Qin, Zhenhua Fan, Zhihang Yu, Zhuo Jiang, and Zijia Wu.
\newblock Minimax-01: Scaling foundation models with lightning attention, 2025.
\newblock URL \url{https://arxiv.org/abs/2501.08313}.

\bibitem[Openai(2024)]{chatml}
Openai.
\newblock Chatml documents, 2024.
\newblock URL \url{https://github.com/openai/openai-python/blob/main/chatml.md}.

\bibitem[OpenAI(2024)]{gpt4o}
OpenAI.
\newblock Hello gpt-4o, 2024.
\newblock URL \url{https://openai.com/index/hello-gpt-4o}.

\bibitem[Ouyang et~al.(2024)Ouyang, Qu, Zhou, Zhu, Zhang, Lin, Wang, Zhao, Jiang, Zhao, Shi, Wu, Chu, Liu, Li, Xu, Zhang, Shi, Tu, and He]{ouyang2024omnidocbenchbenchmarkingdiversepdf}
Linke Ouyang, Yuan Qu, Hongbin Zhou, Jiawei Zhu, Rui Zhang, Qunshu Lin, Bin Wang, Zhiyuan Zhao, Man Jiang, Xiaomeng Zhao, Jin Shi, Fan Wu, Pei Chu, Minghao Liu, Zhenxiang Li, Chao Xu, Bo~Zhang, Botian Shi, Zhongying Tu, and Conghui He.
\newblock Omnidocbench: Benchmarking diverse pdf document parsing with comprehensive annotations, 2024.
\newblock URL \url{https://arxiv.org/abs/2412.07626}.

\bibitem[Paiss et~al.(2023)Paiss, Ephrat, Tov, Zada, Mosseri, Irani, and Dekel]{paiss2023teaching}
Roni Paiss, Ariel Ephrat, Omer Tov, Shiran Zada, Inbar Mosseri, Michal Irani, and Tali Dekel.
\newblock Teaching clip to count to ten.
\newblock In \emph{Proceedings of the IEEE/CVF International Conference on Computer Vision}, pp.\  3170--3180, 2023.

\bibitem[Patraucean et~al.(2024)Patraucean, Smaira, Gupta, Recasens, Markeeva, Banarse, Koppula, Malinowski, Yang, Doersch, et~al.]{patraucean2024perception}
Viorica Patraucean, Lucas Smaira, Ankush Gupta, Adria Recasens, Larisa Markeeva, Dylan Banarse, Skanda Koppula, Mateusz Malinowski, Yi~Yang, Carl Doersch, et~al.
\newblock Perception test: A diagnostic benchmark for multimodal video models.
\newblock In \emph{NeurIPS}, 2024.

\bibitem[Peng et~al.(2023)Peng, Wang, Dong, Hao, Huang, Ma, and Wei]{kosmos2}
Zhiliang Peng, Wenhui Wang, Li~Dong, Yaru Hao, Shaohan Huang, Shuming Ma, and Furu Wei.
\newblock Kosmos-2: Grounding multimodal large language models to the world.
\newblock \emph{arXiv:2306.14824}, 2023.

\bibitem[Rafailov et~al.(2023)Rafailov, Sharma, Mitchell, Manning, Ermon, and Finn]{DBLP:conf/nips/RafailovSMMEF23}
Rafael Rafailov, Archit Sharma, Eric Mitchell, Christopher~D. Manning, Stefano Ermon, and Chelsea Finn.
\newblock Direct preference optimization: Your language model is secretly a reward model.
\newblock In Alice Oh, Tristan Naumann, Amir Globerson, Kate Saenko, Moritz Hardt, and Sergey Levine (eds.), \emph{Advances in Neural Information Processing Systems 36: Annual Conference on Neural Information Processing Systems 2023, NeurIPS 2023, New Orleans, LA, USA, December 10 - 16, 2023}, 2023.
\newblock URL \url{http://papers.nips.cc/paper\_files/paper/2023/hash/a85b405ed65c6477a4fe8302b5e06ce7-Abstract-Conference.html}.

\bibitem[Rawles et~al.(2024)Rawles, Clinckemaillie, Chang, Waltz, Lau, Fair, Li, Bishop, Li, Campbell-Ajala, et~al.]{rawles2024androidworld}
Christopher Rawles, Sarah Clinckemaillie, Yifan Chang, Jonathan Waltz, Gabrielle Lau, Marybeth Fair, Alice Li, William Bishop, Wei Li, Folawiyo Campbell-Ajala, et~al.
\newblock Androidworld: A dynamic benchmarking environment for autonomous agents.
\newblock \emph{arXiv:2405.14573}, 2024.

\bibitem[Rein et~al.(2023)Rein, Hou, Stickland, Petty, Pang, Dirani, Michael, and Bowman]{gpqa}
David Rein, Betty~Li Hou, Asa~Cooper Stickland, Jackson Petty, Richard~Yuanzhe Pang, Julien Dirani, Julian Michael, and Samuel~R. Bowman.
\newblock {GPQA}: A graduate-level {Google}-proof {Q}{\&}{A} benchmark.
\newblock \emph{CoRR}, abs/2311.12022, 2023.

\bibitem[Ren et~al.(2024)Ren, Jiang, Liu, Zeng, Liu, Gao, Huang, Ma, Jiang, Chen, et~al.]{ren2024grounding}
Tianhe Ren, Qing Jiang, Shilong Liu, Zhaoyang Zeng, Wenlong Liu, Han Gao, Hongjie Huang, Zhengyu Ma, Xiaoke Jiang, Yihao Chen, et~al.
\newblock Grounding dino 1.5: Advance the" edge" of open-set object detection.
\newblock \emph{arXiv preprint arXiv:2405.10300}, 2024.

\bibitem[Riquelme et~al.(2021)Riquelme, Puigcerver, Mustafa, Neumann, Jenatton, Susano~Pinto, Keysers, and Houlsby]{riquelme2021scaling}
Carlos Riquelme, Joan Puigcerver, Basil Mustafa, Maxim Neumann, Rodolphe Jenatton, Andr{\'e} Susano~Pinto, Daniel Keysers, and Neil Houlsby.
\newblock Scaling vision with sparse mixture of experts.
\newblock \emph{Advances in Neural Information Processing Systems}, 34:\penalty0 8583--8595, 2021.

\bibitem[Singh et~al.(2019)Singh, Natarajan, Shah, Jiang, Chen, Batra, Parikh, and Rohrbach]{textvqa}
Amanpreet Singh, Vivek Natarajan, Meet Shah, Yu~Jiang, Xinlei Chen, Dhruv Batra, Devi Parikh, and Marcus Rohrbach.
\newblock Towards vqa models that can read.
\newblock In \emph{CVPR}, 2019.

\bibitem[Su et~al.(2024)Su, Ahmed, Lu, Pan, Bo, and Liu]{rope}
Jianlin Su, Murtadha H.~M. Ahmed, Yu~Lu, Shengfeng Pan, Wen Bo, and Yunfeng Liu.
\newblock Roformer: Enhanced {Transformer} with rotary position embedding.
\newblock \emph{Neurocomputing}, 568:\penalty0 127063, 2024.

\bibitem[Tang et~al.(2024)Tang, Liu, Ye, Lu, Wei, Lin, Li, Mahmood, Feng, Zhao, Wang, Liu, Liu, Bai, and Huang]{tang2024mtvqa}
Jingqun Tang, Qi~Liu, Yongjie Ye, Jinghui Lu, Shu Wei, Chunhui Lin, Wanqing Li, Mohamad Fitri Faiz~Bin Mahmood, Hao Feng, Zhen Zhao, Yanjie Wang, Yuliang Liu, Hao Liu, Xiang Bai, and Can Huang.
\newblock Mtvqa: Benchmarking multilingual text-centric visual question answering.
\newblock \emph{arXiv:2405.11985}, 2024.

\bibitem[Team et~al.(2023)Team, Anil, Borgeaud, Wu, Alayrac, Yu, Soricut, Schalkwyk, Dai, Hauth, et~al.]{team2023gemini}
Gemini Team, Rohan Anil, Sebastian Borgeaud, Yonghui Wu, Jean-Baptiste Alayrac, Jiahui Yu, Radu Soricut, Johan Schalkwyk, Andrew~M Dai, Anja Hauth, et~al.
\newblock Gemini: a family of highly capable multimodal models.
\newblock \emph{arXiv preprint arXiv:2312.11805}, 2023.

\bibitem[Tong et~al.(2024)Tong, Brown, Wu, Woo, Middepogu, Akula, Yang, Yang, Iyer, Pan, et~al.]{tong2024cambrian}
Shengbang Tong, Ellis Brown, Penghao Wu, Sanghyun Woo, Manoj Middepogu, Sai~Charitha Akula, Jihan Yang, Shusheng Yang, Adithya Iyer, Xichen Pan, et~al.
\newblock Cambrian-1: A fully open, vision-centric exploration of multimodal llms.
\newblock \emph{arXiv preprint arXiv:2406.16860}, 2024.

\bibitem[Wang et~al.(2024{\natexlab{a}})Wang, Fu, Huang, Li, Liu, Liu, Ma, Xu, Zhou, Zhang, et~al.]{wang2024muirbench}
Fei Wang, Xingyu Fu, James~Y Huang, Zekun Li, Qin Liu, Xiaogeng Liu, Mingyu~Derek Ma, Nan Xu, Wenxuan Zhou, Kai Zhang, et~al.
\newblock Muirbench: A comprehensive benchmark for robust multi-image understanding.
\newblock \emph{arXiv preprint arXiv:2406.09411}, 2024{\natexlab{a}}.

\bibitem[Wang et~al.(2024{\natexlab{b}})Wang, Xu, Jia, Zhang, Yan, Shen, Zhang, Huang, and Sang]{wang2024mobile2}
Junyang Wang, Haiyang Xu, Haitao Jia, Xi~Zhang, Ming Yan, Weizhou Shen, Ji~Zhang, Fei Huang, and Jitao Sang.
\newblock Mobile-agent-v2: Mobile device operation assistant with effective navigation via multi-agent collaboration.
\newblock \emph{arXiv preprint arXiv:2406.01014}, 2024{\natexlab{b}}.

\bibitem[Wang et~al.(2024{\natexlab{c}})Wang, Xu, Ye, Yan, Shen, Zhang, Huang, and Sang]{wang2024mobile}
Junyang Wang, Haiyang Xu, Jiabo Ye, Ming Yan, Weizhou Shen, Ji~Zhang, Fei Huang, and Jitao Sang.
\newblock Mobile-agent: Autonomous multi-modal mobile device agent with visual perception.
\newblock \emph{arXiv preprint arXiv:2401.16158}, 2024{\natexlab{c}}.

\bibitem[Wang et~al.(2024{\natexlab{d}})Wang, Pan, Shi, Lu, Zhan, and Li]{mathvision}
Ke~Wang, Junting Pan, Weikang Shi, Zimu Lu, Mingjie Zhan, and Hongsheng Li.
\newblock Measuring multimodal mathematical reasoning with math-vision dataset.
\newblock \emph{arXiv:2402.14804}, 2024{\natexlab{d}}.

\bibitem[Wang et~al.(2024{\natexlab{e}})Wang, Bai, Tan, Wang, Fan, Bai, Chen, Liu, Wang, Ge, Fan, Dang, Du, Ren, Men, Liu, Zhou, Zhou, and Lin]{Qwen2-VL}
Peng Wang, Shuai Bai, Sinan Tan, Shijie Wang, Zhihao Fan, Jinze Bai, Keqin Chen, Xuejing Liu, Jialin Wang, Wenbin Ge, Yang Fan, Kai Dang, Mengfei Du, Xuancheng Ren, Rui Men, Dayiheng Liu, Chang Zhou, Jingren Zhou, and Junyang Lin.
\newblock Qwen2-vl: Enhancing vision-language model's perception of the world at any resolution.
\newblock \emph{arXiv:2409.12191}, 2024{\natexlab{e}}.

\bibitem[Wang et~al.(2024{\natexlab{f}})Wang, Bai, Tan, Wang, Fan, Bai, Chen, Liu, Wang, Ge, et~al.]{wang2024qwen2}
Peng Wang, Shuai Bai, Sinan Tan, Shijie Wang, Zhihao Fan, Jinze Bai, Keqin Chen, Xuejing Liu, Jialin Wang, Wenbin Ge, et~al.
\newblock Qwen2-vl: Enhancing vision-language model's perception of the world at any resolution.
\newblock \emph{arXiv preprint arXiv:2409.12191}, 2024{\natexlab{f}}.

\bibitem[Wang et~al.(2024{\natexlab{g}})Wang, He, Hong, Cheng, Zhang, Qi, Gu, Huang, Xu, Dong, et~al.]{wang2024lvbench}
Weihan Wang, Zehai He, Wenyi Hong, Yean Cheng, Xiaohan Zhang, Ji~Qi, Xiaotao Gu, Shiyu Huang, Bin Xu, Yuxiao Dong, et~al.
\newblock Lvbench: An extreme long video understanding benchmark.
\newblock \emph{arXiv preprint arXiv:2406.08035}, 2024{\natexlab{g}}.

\bibitem[Wang et~al.(2024{\natexlab{h}})Wang, Ren, Luo, Li, Yan, Chen, Wang, Li, Lu, Zhu, et~al.]{wang2024allseeing_v2}
Weiyun Wang, Yiming Ren, Haowen Luo, Tiantong Li, Chenxiang Yan, Zhe Chen, Wenhai Wang, Qingyun Li, Lewei Lu, Xizhou Zhu, et~al.
\newblock The all-seeing project v2: Towards general relation comprehension of the open world.
\newblock \emph{arXiv preprint arXiv:2402.19474}, 2024{\natexlab{h}}.

\bibitem[Wang et~al.(2023)Wang, Dai, Chen, Huang, Li, Zhu, Hu, Lu, Lu, Li, et~al.]{wang2023internimage}
Wenhai Wang, Jifeng Dai, Zhe Chen, Zhenhang Huang, Zhiqi Li, Xizhou Zhu, Xiaowei Hu, Tong Lu, Lewei Lu, Hongsheng Li, et~al.
\newblock Internimage: Exploring large-scale vision foundation models with deformable convolutions.
\newblock In \emph{Proceedings of the IEEE/CVF conference on computer vision and pattern recognition}, pp.\  14408--14419, 2023.

\bibitem[Wang et~al.(2024{\natexlab{i}})Wang, Zhang, Luo, Sun, Cui, Wang, Zhang, Wang, Li, Yu, et~al.]{wang2024emu3}
Xinlong Wang, Xiaosong Zhang, Zhengxiong Luo, Quan Sun, Yufeng Cui, Jinsheng Wang, Fan Zhang, Yueze Wang, Zhen Li, Qiying Yu, et~al.
\newblock Emu3: Next-token prediction is all you need.
\newblock \emph{arXiv preprint arXiv:2409.18869}, 2024{\natexlab{i}}.

\bibitem[Wang et~al.(2024{\natexlab{j}})Wang, Ma, Zhang, Ni, Chandra, Guo, Ren, Arulraj, He, Jiang, Li, Ku, Wang, Zhuang, Fan, Yue, and Chen]{mmlupro}
Yubo Wang, Xueguang Ma, Ge~Zhang, Yuansheng Ni, Abhranil Chandra, Shiguang Guo, Weiming Ren, Aaran Arulraj, Xuan He, Ziyan Jiang, Tianle Li, Max Ku, Kai Wang, Alex Zhuang, Rongqi Fan, Xiang Yue, and Wenhu Chen.
\newblock {MMLU-Pro}: {A} more robust and challenging multi-task language understanding benchmark.
\newblock \emph{CoRR}, abs/2406.01574, 2024{\natexlab{j}}.

\bibitem[Wang et~al.(2025)Wang, Xu, Wang, Zhang, Yan, Zhang, Huang, and Ji]{wang2025mobile}
Zhenhailong Wang, Haiyang Xu, Junyang Wang, Xi~Zhang, Ming Yan, Ji~Zhang, Fei Huang, and Heng Ji.
\newblock Mobile-agent-e: Self-evolving mobile assistant for complex tasks.
\newblock \emph{arXiv preprint arXiv:2501.11733}, 2025.

\bibitem[Wang et~al.(2024{\natexlab{k}})Wang, Xia, He, Chen, Liu, Zhu, Liang, Wu, Liu, Malladi, Chevalier, Arora, and Chen]{wang2024charxiv}
Zirui Wang, Mengzhou Xia, Luxi He, Howard Chen, Yitao Liu, Richard Zhu, Kaiqu Liang, Xindi Wu, Haotian Liu, Sadhika Malladi, Alexis Chevalier, Sanjeev Arora, and Danqi Chen.
\newblock Charxiv: Charting gaps in realistic chart understanding in multimodal llms.
\newblock \emph{arXiv preprint arXiv:2406.18521}, 2024{\natexlab{k}}.

\bibitem[Wei et~al.(2022)Wei, Wang, Schuurmans, Bosma, Chi, Le, and Zhou]{DBLP:journals/corr/abs-2201-11903}
Jason Wei, Xuezhi Wang, Dale Schuurmans, Maarten Bosma, Ed~H. Chi, Quoc Le, and Denny Zhou.
\newblock Chain of thought prompting elicits reasoning in large language models.
\newblock \emph{CoRR}, abs/2201.11903, 2022.
\newblock URL \url{https://arxiv.org/abs/2201.11903}.

\bibitem[White et~al.(2024)White, Dooley, Roberts, Pal, Feuer, Jain, Shwartz{-}Ziv, Jain, Saifullah, Naidu, Hegde, LeCun, Goldstein, Neiswanger, and Goldblum]{livebench}
Colin White, Samuel Dooley, Manley Roberts, Arka Pal, Benjamin Feuer, Siddhartha Jain, Ravid Shwartz{-}Ziv, Neel Jain, Khalid Saifullah, Siddartha Naidu, Chinmay Hegde, Yann LeCun, Tom Goldstein, Willie Neiswanger, and Micah Goldblum.
\newblock {LiveBench}: A challenging, contamination-free {LLM} benchmark.
\newblock \emph{CoRR}, abs/2406.19314, 2024.

\bibitem[Wu et~al.(2024{\natexlab{a}})Wu, Li, Chen, and Li]{wu2024longvideobench}
Haoning Wu, Dongxu Li, Bei Chen, and Junnan Li.
\newblock Longvideobench: A benchmark for long-context interleaved video-language understanding, 2024{\natexlab{a}}.
\newblock URL \url{https://arxiv.org/abs/2407.15754}.

\bibitem[Wu et~al.(2024{\natexlab{b}})Wu, Chen, Pan, Liu, Liu, Dai, Gao, Ma, Wu, Wang, et~al.]{wu2024deepseek}
Zhiyu Wu, Xiaokang Chen, Zizheng Pan, Xingchao Liu, Wen Liu, Damai Dai, Huazuo Gao, Yiyang Ma, Chengyue Wu, Bingxuan Wang, et~al.
\newblock Deepseek-vl2: Mixture-of-experts vision-language models for advanced multimodal understanding.
\newblock \emph{arXiv preprint arXiv:2412.10302}, 2024{\natexlab{b}}.

\bibitem[{X.AI}(2024)]{grok15}
{X.AI}.
\newblock Grok-1.5 vision preview.
\newblock \url{https://x.ai/blog/grok-1.5v}, 2024.

\bibitem[Xiao et~al.(2023)Xiao, Wu, Xu, Dai, Hu, Lu, Zeng, Liu, and Yuan]{xiao2023florence}
Bin Xiao, Haiping Wu, Weijian Xu, Xiyang Dai, Houdong Hu, Yumao Lu, Michael Zeng, Ce~Liu, and Lu~Yuan.
\newblock Florence-2: Advancing a unified representation for a variety of vision tasks (2023).
\newblock \emph{URL https://arxiv. org/abs/2311.06242}, 2023.

\bibitem[Xie et~al.(2025)Xie, Zhang, Chen, Li, Zhao, Cao, Toh, Cheng, Shin, Lei, et~al.]{xie2025osworld}
Tianbao Xie, Danyang Zhang, Jixuan Chen, Xiaochuan Li, Siheng Zhao, Ruisheng Cao, Jing~Hua Toh, Zhoujun Cheng, Dongchan Shin, Fangyu Lei, et~al.
\newblock Osworld: Benchmarking multimodal agents for open-ended tasks in real computer environments.
\newblock \emph{Advances in Neural Information Processing Systems}, 37:\penalty0 52040--52094, 2025.

\bibitem[Xu et~al.(2024)Xu, Wang, Wang, Lu, Xie, Saha, Sahoo, Yu, and Xiong]{xu2024aguvis}
Yiheng Xu, Zekun Wang, Junli Wang, Dunjie Lu, Tianbao Xie, Amrita Saha, Doyen Sahoo, Tao Yu, and Caiming Xiong.
\newblock Aguvis: Unified pure vision agents for autonomous gui interaction.
\newblock \emph{arXiv preprint arXiv:2412.04454}, 2024.

\bibitem[Yang et~al.(2024{\natexlab{a}})Yang, Yang, Zhang, Hui, Zheng, Yu, Li, Liu, Huang, et~al.]{qwen2.5}
An~Yang, Baosong Yang, Beichen Zhang, Binyuan Hui, Bo~Zheng, Bowen Yu, Chengyuan Li, Dayiheng Liu, Fei Huang, et~al.
\newblock Qwen2.5 technical report.
\newblock \emph{arXiv:2412.15115}, 2024{\natexlab{a}}.

\bibitem[Yang et~al.(2024{\natexlab{b}})Yang, Tang, Li, Wang, Wan, Zhong, Liu, Yang, Wang, Bai, Jin, and Lin]{yang2024ccocrcomprehensivechallengingocr}
Zhibo Yang, Jun Tang, Zhaohai Li, Pengfei Wang, Jianqiang Wan, Humen Zhong, Xuejing Liu, Mingkun Yang, Peng Wang, Shuai Bai, LianWen Jin, and Junyang Lin.
\newblock Cc-ocr: A comprehensive and challenging ocr benchmark for evaluating large multimodal models in literacy, 2024{\natexlab{b}}.
\newblock URL \url{https://arxiv.org/abs/2412.02210}.

\bibitem[Ye et~al.(2024)Ye, Huang, Lu, Yu, Ping, Tao, Kautz, Han, Xu, Molchanov, et~al.]{ye2024x}
Hanrong Ye, De-An Huang, Yao Lu, Zhiding Yu, Wei Ping, Andrew Tao, Jan Kautz, Song Han, Dan Xu, Pavlo Molchanov, et~al.
\newblock X-vila: Cross-modality alignment for large language model.
\newblock \emph{arXiv preprint arXiv:2405.19335}, 2024.

\bibitem[Ye et~al.(2023)Ye, Xu, Ye, Yan, Liu, Qian, Zhang, Huang, and Zhou]{mplug-owl2}
Qinghao Ye, Haiyang Xu, Jiabo Ye, Ming Yan, Haowei Liu, Qi~Qian, Ji~Zhang, Fei Huang, and Jingren Zhou.
\newblock mplug-owl2: Revolutionizing multi-modal large language model with modality collaboration.
\newblock \emph{arXiv:2311.04257}, 2023.

\bibitem[Yu et~al.(2024)Yu, Yang, Li, Wang, Lin, Liu, Wang, and Wang]{yu2024mm}
Weihao Yu, Zhengyuan Yang, Linjie Li, Jianfeng Wang, Kevin Lin, Zicheng Liu, Xinchao Wang, and Lijuan Wang.
\newblock Mm-vet: Evaluating large multimodal models for integrated capabilities.
\newblock In \emph{ICML}, 2024.

\bibitem[Yue et~al.(2023)Yue, Ni, Zhang, Zheng, Liu, Zhang, Stevens, Jiang, Ren, Sun, et~al.]{yue2023mmmu}
Xiang Yue, Yuansheng Ni, Kai Zhang, Tianyu Zheng, Ruoqi Liu, Ge~Zhang, Samuel Stevens, Dongfu Jiang, Weiming Ren, Yuxuan Sun, et~al.
\newblock Mmmu: A massive multi-discipline multimodal understanding and reasoning benchmark for expert agi.
\newblock \emph{arXiv:2311.16502}, 2023.

\bibitem[Yue et~al.(2024)Yue, Zheng, Ni, Wang, Zhang, Tong, Sun, Yin, Yu, Zhang, et~al.]{mmmupro}
Xiang Yue, Tianyu Zheng, Yuansheng Ni, Yubo Wang, Kai Zhang, Shengbang Tong, Yuxuan Sun, Ming Yin, Botao Yu, Ge~Zhang, et~al.
\newblock Mmmu-pro: A more robust multi-discipline multimodal understanding benchmark.
\newblock \emph{arXiv preprint arXiv:2409.02813}, 2024.

\bibitem[Zhang \& Sennrich(2019)Zhang and Sennrich]{rmsnorm}
Biao Zhang and Rico Sennrich.
\newblock Root mean square layer normalization.
\newblock In \emph{NeurIPS}, 2019.

\bibitem[Zhang et~al.(2024{\natexlab{a}})Zhang, You, Dufter, Zhang, Chen, Chen, Fu, Wang, Chang, Gan, and Yang]{ferretv2}
Haotian Zhang, Haoxuan You, Philipp Dufter, Bowen Zhang, Chen Chen, Hong{-}You Chen, Tsu{-}Jui Fu, William~Yang Wang, Shih{-}Fu Chang, Zhe Gan, and Yinfei Yang.
\newblock Ferret-v2: An improved baseline for referring and grounding with large language models.
\newblock \emph{arXiv:2404.07973}, 2024{\natexlab{a}}.

\bibitem[Zhang et~al.(2024{\natexlab{b}})Zhang, Dong, Cao, Zang, Qian, Wei, Chen, Li, Niu, Ding, et~al.]{zhang2024internlm}
Pan Zhang, Xiaoyi Dong, Yuhang Cao, Yuhang Zang, Rui Qian, Xilin Wei, Lin Chen, Yifei Li, Junbo Niu, Shuangrui Ding, et~al.
\newblock Internlm-xcomposer2. 5-omnilive: A comprehensive multimodal system for long-term streaming video and audio interactions.
\newblock \emph{arXiv preprint arXiv:2412.09596}, 2024{\natexlab{b}}.

\bibitem[Zhang et~al.(2024{\natexlab{c}})Zhang, Jiang, Zhang, Lin, Guo, Qiu, Zhou, Lu, Chang, Qiao, et~al.]{zhang2024mathverse}
Renrui Zhang, Dongzhi Jiang, Yichi Zhang, Haokun Lin, Ziyu Guo, Pengshuo Qiu, Aojun Zhou, Pan Lu, Kai-Wei Chang, Yu~Qiao, et~al.
\newblock Mathverse: Does your multi-modal llm truly see the diagrams in visual math problems?
\newblock In \emph{European Conference on Computer Vision}, pp.\  169--186. Springer, 2024{\natexlab{c}}.

\bibitem[Zhang et~al.(2024{\natexlab{d}})Zhang, Li, Fei, Yuan, Wu, Ji, Loy, and Yan]{zhang2024omg}
Tao Zhang, Xiangtai Li, Hao Fei, Haobo Yuan, Shengqiong Wu, Shunping Ji, Chen~Change Loy, and Shuicheng Yan.
\newblock Omg-llava: Bridging image-level, object-level, pixel-level reasoning and understanding.
\newblock \emph{arXiv preprint arXiv:2406.19389}, 2024{\natexlab{d}}.

\bibitem[Zhang et~al.(2024{\natexlab{e}})Zhang, Wang, Li, Zhang, Taslakian, Rajeswar, Fu, Liu, and Bengio]{zhang2024vcr}
Tianyu Zhang, Suyuchen Wang, Lu~Li, Ge~Zhang, Perouz Taslakian, Sai Rajeswar, Jie Fu, Bang Liu, and Yoshua Bengio.
\newblock Vcr: Visual caption restoration.
\newblock \emph{arXiv:2406.06462}, 2024{\natexlab{e}}.

\bibitem[Zhang et~al.(2024{\natexlab{f}})Zhang, Zhang, Tian, Fu, Zhang, Wu, Li, Wang, Wen, Zhang, et~al.]{mme-realworld}
Yi-Fan Zhang, Huanyu Zhang, Haochen Tian, Chaoyou Fu, Shuangqing Zhang, Junfei Wu, Feng Li, Kun Wang, Qingsong Wen, Zhang Zhang, et~al.
\newblock Mme-realworld: Could your multimodal llm challenge high-resolution real-world scenarios that are difficult for humans?
\newblock \emph{arXiv preprint arXiv:2408.13257}, 2024{\natexlab{f}}.

\bibitem[Zhao et~al.(2025)Zhao, Xie, Zhang, Gan, Long, Hu, Hu, Chen, Li, Song, Xu, Wang, Pan, Shangguan, Tang, Liang, Liu, Zhao, and Cohan]{zhao2025mmvu}
Yilun Zhao, Lujing Xie, Haowei Zhang, Guo Gan, Yitao Long, Zhiyuan Hu, Tongyan Hu, Weiyuan Chen, Chuhan Li, Junyang Song, Zhijian Xu, Chengye Wang, Weifeng Pan, Ziyao Shangguan, Xiangru Tang, Zhenwen Liang, Yixin Liu, Chen Zhao, and Arman Cohan.
\newblock Mmvu: Measuring expert-level multi-discipline video understanding, 2025.
\newblock URL \url{https://arxiv.org/abs/2501.12380}.

\bibitem[Zhou et~al.(2023)Zhou, Lu, Mishra, Brahma, Basu, Luan, Zhou, and Hou]{ifeval}
Jeffrey Zhou, Tianjian Lu, Swaroop Mishra, Siddhartha Brahma, Sujoy Basu, Yi~Luan, Denny Zhou, and Le~Hou.
\newblock Instruction-following evaluation for large language models.
\newblock \emph{CoRR}, abs/2311.07911, 2023.

\bibitem[Zhou et~al.(2024)Zhou, Shu, Zhao, Wu, Xiao, Yang, Xiong, Zhang, Huang, and Liu]{zhou2024mlvu}
Junjie Zhou, Yan Shu, Bo~Zhao, Boya Wu, Shitao Xiao, Xi~Yang, Yongping Xiong, Bo~Zhang, Tiejun Huang, and Zheng Liu.
\newblock Mlvu: A comprehensive benchmark for multi-task long video understanding.
\newblock \emph{arXiv preprint arXiv:2406.04264}, 2024.

\end{thebibliography}
\bibliographystyle{colm2024_conference}


\end{document}